\documentclass[runningheads]{llncs}
\usepackage{graphicx}
\usepackage{amsmath,amssymb} %
\usepackage{color}
\usepackage{url}
\usepackage{scrextend}
\usepackage{hyperref}

\usepackage{footnote}
\usepackage{multirow}
\graphicspath{{./im/}}
\newcommand\NB{IconArt} %
\newcommand\MIL{MI-max}
\newcommand\MILS{MI-max}
\newcommand\MILSCV{MI-max-C}
\newcommand\MAX{MAX}
\newcommand{\heightimage}{2.55cm}
\newcommand{\repeatfootnote}{\textsuperscript{\thefootnote}}

\begin{document}
\title{Weakly Supervised Object Detection in Artworks} 

\titlerunning{Weakly Supervised Detection in Artworks}
\author{Nicolas Gonthier\inst{1} \and
Yann Gousseau\inst{1} \and
Said Ladjal\inst{1} \and
Olivier Bonfait\inst{2}}
\authorrunning{N. Gonthier et al.}

\institute{LTCI, Telecom ParisTech, Universite Paris-Saclay, 75013, Paris, France \\
\email{\{firstname.lastname\}@telecom-paristech.fr}\\
\and
Universite de Bourgogne, UMR CNRS UB 5605, 21000, Dijon, France\\
}

\maketitle              

\begin{abstract}
We propose a method for the weakly supervised detection of objects in paintings. At training time, only image-level annotations are needed. This, combined with the efficiency of our multiple-instance learning method, enables one to learn new classes on-the-fly from globally annotated databases, avoiding the tedious task of manually marking objects. We show on several databases that dropping the instance-level annotations only yields mild performance losses. We also introduce a new database, IconArt, on which we perform detection experiments on classes that could not be learned on photographs, such as Jesus Child or Saint Sebastian. To the best of our knowledge, these are the first experiments dealing with the automatic (and in our case weakly supervised) detection of iconographic elements in paintings. We believe that such a method is of great benefit for helping art historians to explore large digital databases. 
\keywords{weakly supervised detection  \and transfer learning \and  art analysis \and multiple instance learning}
\end{abstract}

\section{Introduction}
Several recent works show that recycling analysis tools that have been developed for natural images (photographs) can yield surprisingly good results for analysing paintings or drawings. In particular, impressive classification results are obtained on painting databases by using convolutional neural networks (CNNs) designed for the classification of photographs~\cite{crowley_search_2014,yin2016object}. These results occur in a general context were methods of transfer learning~\cite{donahue_decaf_2013} (changing the task a model was trained for) and domain adaptation (changing the nature of the data a model was trained on) are increasingly applied. Classifying and analysing paintings is of course of great interest to art historians, and can help them to take full advantage of the massive artworks databases that are built worldwide. 

More difficult than classification, and at the core of many recent computer vision works, the object detection task (classifying and localising an object) has been less studied in the case of paintings, although exciting results have been obtained, again using transfer techniques~\cite{crowley_art_2016,westlake_detecting_2016,inoue_crossdomain_2018}. 

Methods that detect objects in photographs have been developed thanks to massive image databases on which several classes (such as cats, people, cars) have been manually localised with bounding boxes. The PASCAL VOC~\cite{everingham_pascal_2007} and MS COCO~\cite{lin_microsoft_2014} datasets have been crucial in the development of detection methods and the recently introduced Google Open Image Dataset (2M images, 15M boxes for 600 classes) is expected to push further the limits of detection. Now, there is no such database (with localised objects) in the field of Art History, even though large databases are being build by many institutions or academic research teams, e.g.~\cite{rijksmuseum_online_2018,rmn_online,europeana_collections_2018,met_image_2018,pharosconsortium_pharos_2018,wilber_bam_2017}. Some of these databases include image-level annotations, but none includes location annotations. Besides, manually annotating such large databases is tedious and must be performed each time a new category is searched for. Therefore, it is of great interest to develop {\it weakly supervised} detection methods, that can learn to detect objects using image-level annotations only. While this aspect has been thoroughly studied for natural images, only a few studies have been dedicated to the case of painting or drawings.

Moreover, these studies are mostly dedicated to the cross depiction problem: they learn to detect the same objects in photographs and in paintings, in particular man-made objects (cars, bottles ...) or animals. While these may be useful to art historians, it is obviously needed to detect more specific objects or attributes such as ruins or nudity, and characters of iconographic interest such as Mary, Jesus as a child or the crucifixion of Jesus, for instance. These last categories can hardly be directly inherited from photographic databases. 

For these two reasons, the lack of location annotations and the specificity of the categories of interest, a general method allowing the weakly supervised detection on specific domains such as paintings would be of great interest to art historians and more generally to anyone needing some automatic tools to explore artistic databases. We propose some contributions in this direction:
\begin{itemize}
    \item We introduce a new multiple-instance learning (MIL) technique that is simple and quick enough to deal with large databases,
\item We demonstrate the utility of the proposed technique for object detection on weakly annotated databases, including photographs, drawings and paintings. These experiments are performed using image-level annotations only. 
\item We propose the first experiments dealing with the recognition and detection of iconographic elements that are specific to Art History, exhibiting both successful detections and some classes that are particularly challenging, especially in a weakly supervised context.

\end{itemize}

We believe that such a system, enabling one to detect new and unseen category with minimal supervision, is of great benefit for dealing efficiently with digital artwork databases. More precisely, iconographic detection results are useful for different and particularly active domains of humanities: Art History (to gather data relative to the iconography of recurrent characters, such as the Virgin Mary or San Sebastian, as well as to study the formal evolution of their representations), Semiology (to infer mutual configurations or relative dimensions of the iconographic elements), History of Ideas and Cultures (with category such as nudity, ruins), Material Culture Studies, etc. 

In particular, being able to detect iconographic elements is of great importance for the study of spatial configurations, which are central to the reading of images and particularly timely given the increasing importance of Semiology. To fix ideas, we can give two examples of potential use. First, the order in which iconographic elements are encountered (e.g. Gabriel and Mary), when reading an image from left to right, has received much attention from art historians~\cite{gasparro2008dal}. In the same spirit, recent studies~\cite{deBosio2017} on the meaning of mirror images in early modern Italy could benefit from the detection of iconographic elements.

\section{Related Work}

\textbf{Object recognition and detection in artworks}
Early works on cross-domain (or cross-depiction) image comparisons were mostly concerned with sketch retrieval, see e.g.~\cite{del1997visual}. Various local descriptors were then used for comparing and classifying images, such as
part-based models \cite{shrivastava2011data} or mid-level discriminative patches \cite{aubry2014painting,crowley2014state}. In order to enhance the generalisation capacity of these approaches, it was proposed in~\cite{wu2014learning} to model object through graphs of labels. More generally, it was shown in~\cite{hall2015cross} that structured models are more prone to succeed in cross-domain recognition than appearance-based models. 

Next, several works have tried to transfer the tremendous classification capacity of convolutional neural networks to perform cross-domain object recognition, in particular for paintings. In~\cite{crowley_search_2014}, it is shown that recycling CNNs directly for the task of recognising objects in paintings, without fine-tuning, yields surprisingly good results. Similar conclusions were also given in~\cite{yin2016object} for artistic drawings. In~\cite{li_deeper_2017}, a robust low rank parametrized CNN model is proposed to recognise common categories in an unseen domain (photo, painting, cartoon or sketch).
In~\cite{wilber_bam_2017}, a new annotated database is introduced, on which it is shown that fine-tuning improves recognition performances. Several works have also successfully adapted CNNs architectures to the problem of style recognition in artworks~\cite{lecoutre_recognizing_2017,bianco2017deep,mao_deepart_2017}. More generally, the use of CNNs opens the way to other artwork analysis tasks, such as  visual links retrieval~\cite{seguin_visual_2016}, scene classification~\cite{florea_domain_2017}, author classification~\cite{vannoord_learning_2017} or possibly to generic artwork content representation~\cite{strezoski_omniart_2017}. 

The problem of {\it object detection} in paintings, that is, being able to both localise and recognise objects, has been less studied. In~\cite{crowley_art_2016}, it is shown that applying a pre-trained object detector (Faster R-CNN~\cite{ren_faster_2015}) and then selecting the localisation with highest confidence can yield correct detections of PASCAL VOC classes. Other works attacked this difficult problem by restricting it to a single class. In~\cite{ginosar2014detecting}, it is shown that deformable part model  outperforms other approaches, including some CNNs, for the detection of people in cubist artworks. In~\cite{redmon_you_2016}, it is shown that the YOLO network trained on natural images can, to some extend, be used for people detection in cubism.
In~\cite{westlake_detecting_2016}, it is proposed to perform people detection in a wide variety of artworks (through a newly introduced database) by fine-tuning a network in a supervised way. People can be detected with high accuracy even though the database has very large stylistic variations and includes paintings that strongly differs from photographs in the way they represent people.

\textbf{Weakly supervised detection} refers to the task of learning an object detector using limited annotations, usually image-level annotations only. Often, a set of detections (e.g. bounding boxes) is considered at image level, assuming we only know if at least one of the detection corresponds the category of interest. The corresponding statistical problem is referred to as multiple instance learning (MIL) \cite{dietterich1997solving}. A well-known solution to this problem through a generalisation of Support Vector Machine (SVM) has been proposed in \cite{andrews_support_2003}. Several approximations of the involved non-convex problem have been proposed, see e.g. \cite{gehler2007deterministic} or the recent survey~\cite{carbonneau_multiple_2016}.

Recently, this problem has been attacked using classification and detection neural networks. In \cite{song_learning_2014}, it is proposed to learn a smooth version of an SVM  on the features from R-CNN \cite{girshick_rich_2014} and to focus on the initialisation phase which is crucial due to the non-convexity of the problem.
In \cite{redmon2016}, it is proposed to learn to detect new specific classes by taking advantage of the knowledge of wider classes. 
In \cite{bilen_weakly_2016} a weakly supervised deep detection network is proposed based on Fast R-CNN \cite{girshick_fast_2015}. Those works have been improved in \cite{tang_multiple_2017} by adding a multi-stage classifier refinement.
In \cite{cinbis_weakly_2016} a multi-fold split of the training data is proposed to escape local optima. 
In \cite{li_weakly_2016}, a two step strategy is proposed, first collecting good regions by a mask-out classification, then selecting the best positive region in each image by a MIL formulation and then fine-tuning a detector with those propositions as "ground truth" bounding boxes. 
In \cite{durand_wildcat_2017} a new pooling strategy is proposed to efficiently learn localisation of objects without doing bounding boxes regression.

Weakly supervised strategies for the cross domain problem have been much less studied. In~\cite{crowley_art_2016}, a relatively basic methodology is proposed, in which for each image the bounding box with highest (class agnostic) "objectness" score is classified.
In \cite{inoue_crossdomain_2018}, it is proposed to do mixed supervised object detection with cross-domain learning based on the SSD network \cite{liu2016}. Object detectors are learnt by using instance-level annotations on photographs and image-level annotations on a target domain (watercolor, cartoon, etc.). We will perform comparisons of our approach with these two methods in Section~\ref{sec:experiments}.

\section{Weakly supervised detection by transfer learning}

In this section, we propose our approach to the weakly supervised detection of visual category in paintings. In order to perform transfer learning, we first apply Faster R-CNN~\cite{ren_faster_2015} (a detection network trained on photographs) which is used as a feature extractor, in the same way as in~\cite{crowley_art_2016}. This results in a set of candidate bounding boxes. For a given visual category, the goal is then, using image-level annotations only, to decide which boxes correspond to this category. For this, we propose a new multiple-instance learning method, that will be detailed in Section~\ref{MILsection}. In contrast with classical approaches to the MIL problem such as~\cite{andrews_support_2003} the proposed heuristic is very fast. This, combined with the fact that we do not need fine-tuning, permits a flexible on-the-fly learning of new category in a few minutes. 

\begin{figure}
    \centering
  \includegraphics[height=4.5cm]{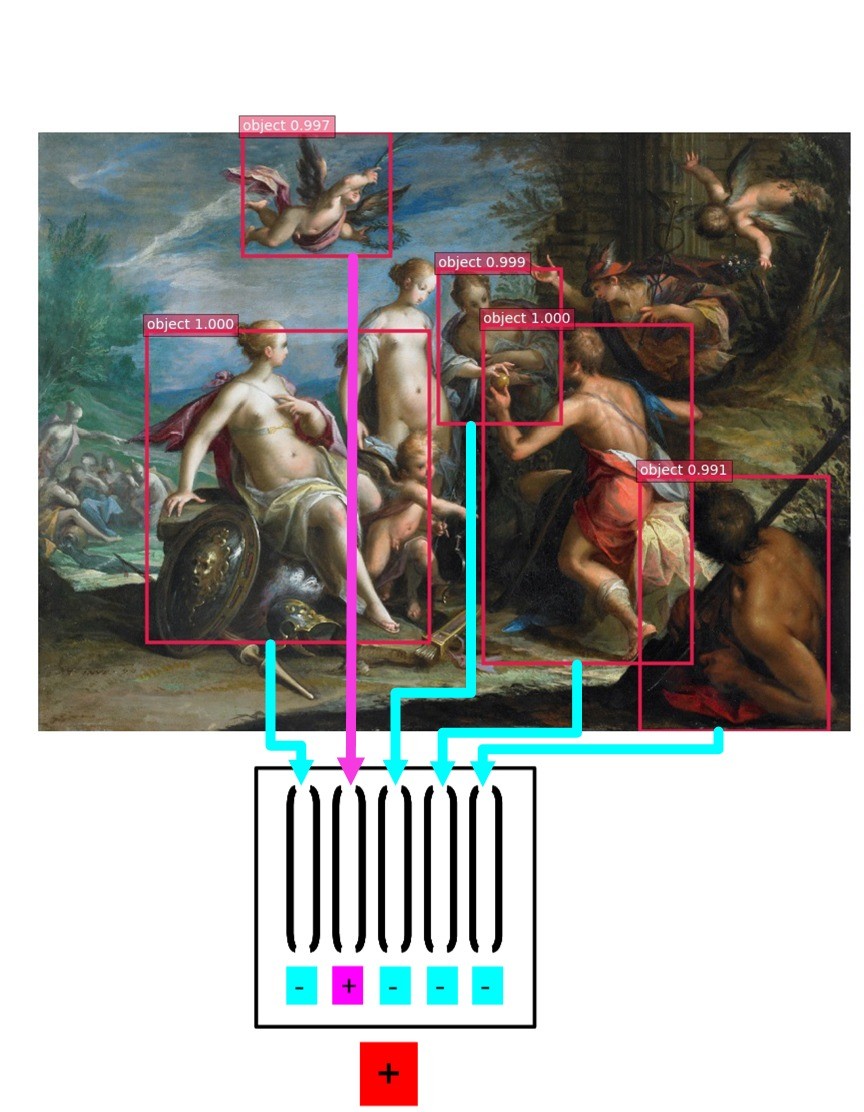} \hspace{2cm}
\includegraphics[height=4.5cm]{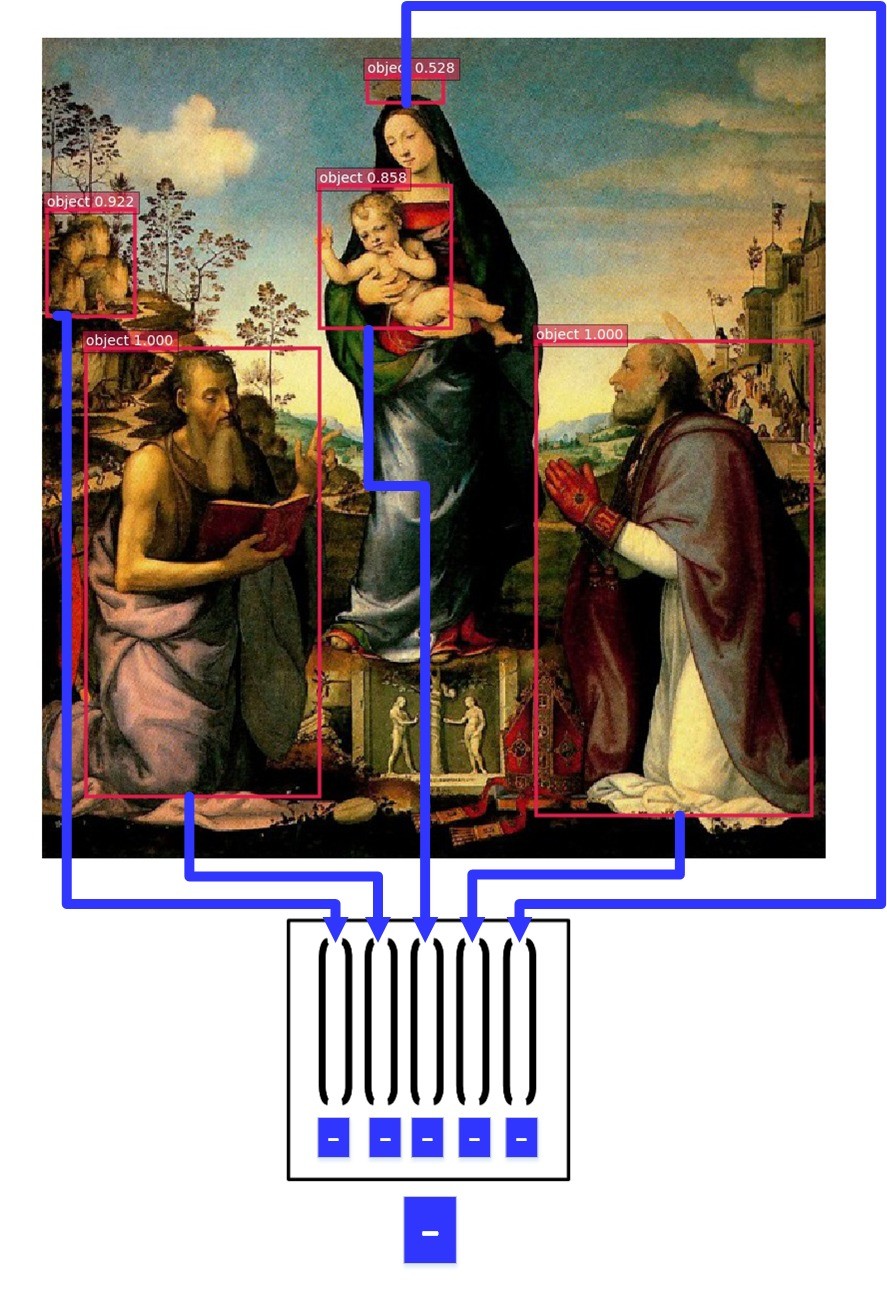}
    \caption{Illustration of positive and negative sets of detections (bounding boxes) for the {\it angel} category.}
    \label{fig:angel-Bag}

\end{figure}

Figure~\ref{fig:angel-Bag} illustrates the situation we face at training time. For each image, we are given a set of bounding boxes which receive a label $+1$ (the visual category of interest is present at least once) or $-1$ (the category is not present in this image). 

\subsection{Multiple Instance Learning}
\label{MILsection}

 The usual way to perform MIL is through the resolution of a non-convex energy minimisation~\cite{andrews_support_2003}, although efficient convex relaxations have been proposed~\cite{joulin2012convex}. One disadvantage of these approaches is their heavy computational cost. In what follows, we propose a simple and fast heuristic to this problem.

For simplicity of the presentation, we assume only one visual category. Assume we have $N$ images at hand, each of which contains $K$ bounding boxes. Each image receives a label $y = +1$ when it is a positive example (the category is present) and $y = -1$ otherwise. We denote by $n_1$ the number of positive examples  in the training set, and by $n_{-1}$ the number of negative examples.

Images are indexed by $i$, the $K$ regions provided by the object detector are indexed by $k$, the label of the $i$-th image is denoted by $y_i$ and the high level semantic feature vector of size M associated to the $k$-th box in the $i$-th image is denoted $X_{i,k}$. We also assume that the detector provides a (class agnostic) "objectness" score for this box, denoted $s_{i,k}$.

We make the (strong) hypothesis that if $y_{i} = +1 $, then there is at least one of the $K$ regions in image $i$ that contains an occurrence of the category. In a sense, we assume that the region proposal part is robust enough to transfer detections from photography to the target domain.

Following this assumption, our problem boils down to the classic multiple-instance classification problem~\cite{dietterich1997solving}: if for image $i$ we have $y_i=+1$, then at least one of the boxes contains the category, whereas if $y_i=-1$ no box does. The goal is then to decide which boxes correspond to the category. Instead of the classical SVM generalisation proposed in \cite{andrews_support_2003} and based on an iterative procedure, we look for an hyperplan minimising the functional defined below. We look for $w \in \mathbf{R}^M$, $b\in\mathbf{R}$ achieving

\begin{equation}
  min_{(w,b)}  \mathcal{L} (w,b) 
\end{equation}
 with 

\begin{equation}
\phi(w,b) = \sum_{i=1}^{N} \frac{-y_{i}}{n_{y_i}} Tanh \left\lbrace \max_{ k \in \{1..K \} }  \left(  w^{T} X_{i,k} + b \right) \right\rbrace
\end{equation}

and 
\begin{equation}
\label{eqn:LossReg}
\mathcal{L} (w,b) =\phi(w,b) + C * ||w||^2,
\end{equation}
where $C$ is a constant balancing the regularisation term. 
The intuition behind this formulation is that minimising $\mathcal{L} (w,b) $ amounts to seek a hyperplan separating the most positive element of each positive image from the least negative element of the negative image, sharing similar ideas as in
MI-SVM \cite{andrews_support_2003} or Latent-SVM \cite{felzenszwalb_object_2010}. The $\mathop{Tanh}$ is here to mimic the SVM formulation in which only the worst margins count. 
We divide by $n_{y_i}$ to account for unbalanced data. Indeed most example images are negative ones ($n_{-1}>>n_{1})$).

The main advantage of this formulation is that it can be realised by a simple gradient descent, therefore avoiding costly multiple SVM optimisation.
If the dataset is too big to fit in the memory, we switch to a stochastic gradient descent by considering random batches in the training set.

As this problem is non-convex, we try several random initialisation and we select the couple $w,b$ minimising the classification function $\phi(w,b)$. Although we did not explore this possibility it may be interesting to keep more than one vector to describe a class, since one iconographic element could have more that one specific feature, each stemming from a distinctive part.

In practice, we observed consistently better results when modifying slightly the above formulation by considering the (class-agnostic) "objectness" score associated to each box (as returned by Faster R-CNN). Therefore we modify function $\phi$ to 
\begin{equation}
\label{eqn:MILS}
\phi^s(w,b) = \sum_{i=1}^{N} \frac{-y_{i}}{n_{y_i}} Tanh \left\lbrace \max_{k \in \{1..K\} } \left(  \left( s_{i,k} + \epsilon \right) \left(w^{T} X_{i,k} + b \right)  \right) \right\rbrace
\end{equation}
with $ \epsilon \geq 0$. 
The motivation behind this formulation is that the score $s_{i,k}$, roughly a probability that there is an object (of any category) in box $k$, provides a prioritisation between boxes.

Once the best couple ($w^{\star},b^{\star}$) has been found, we compute the following score, reflecting the meaningfulness of category association :
\begin{equation}
    S(x) = Tanh  \lbrace \left( s(x) + \epsilon \right) \left(w^{\star T} x + b^{\star} \right) \rbrace
    \label{eq:scoreMILS}
\end{equation}

At test time, each box with a positive score \eqref{eq:scoreMILS} (where $s(x)$ is the objectness score associated to $x$) is affected to the category. The approach is then straightforwardly extended to an arbitrary number of categories, by computing a couple $(w^{\star},b^{\star})$ per category. Observe, however, that this leads to non-comparable scores between categories. 
Among all boxes affected to each class, a non-maximal suppression (NMS) algorithm is then applied in order to avoid redundant detections. The resulting multiple instance learning method is called {\bf \MIL{}}. 

\subsection{Implementation details}
\label{sec:implementation}

\paragraph{\bf Faster R-CNN} We use the detection network Faster R-CNN~\cite{ren_faster_2015}. We only keep its region proposal part (RPN) and the features corresponding to each proposed region. In order to yield and efficient and flexible learning of new classes, we choose to avoid  retraining or even fine-tuning. 
Faster R-CNN is a meta-network in which a pre-trained  network is enclosed. The quality of features depends on the enclosed network and we compare several possibility in the experimental part. 

\begin{figure}[!tbp] 
  \centering
  \hfill
    \includegraphics[height=5cm]{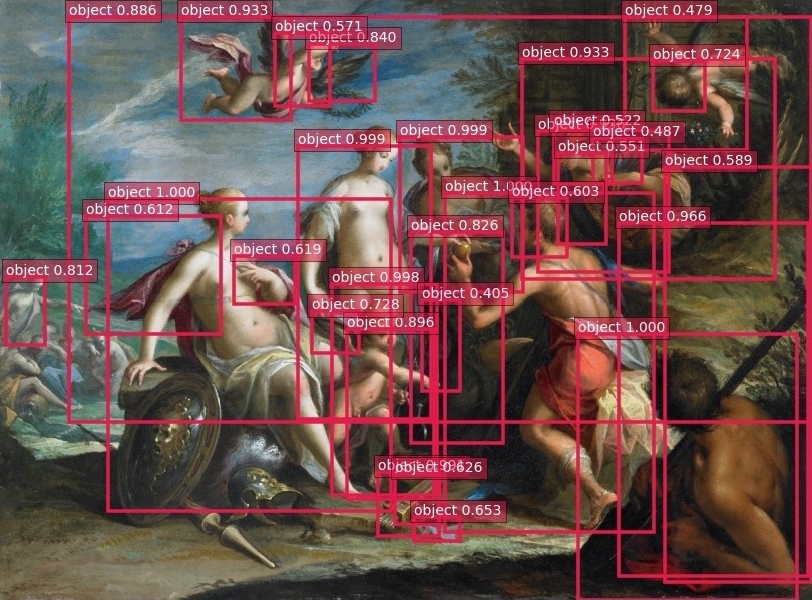} \hfill
    \includegraphics[height=5cm]{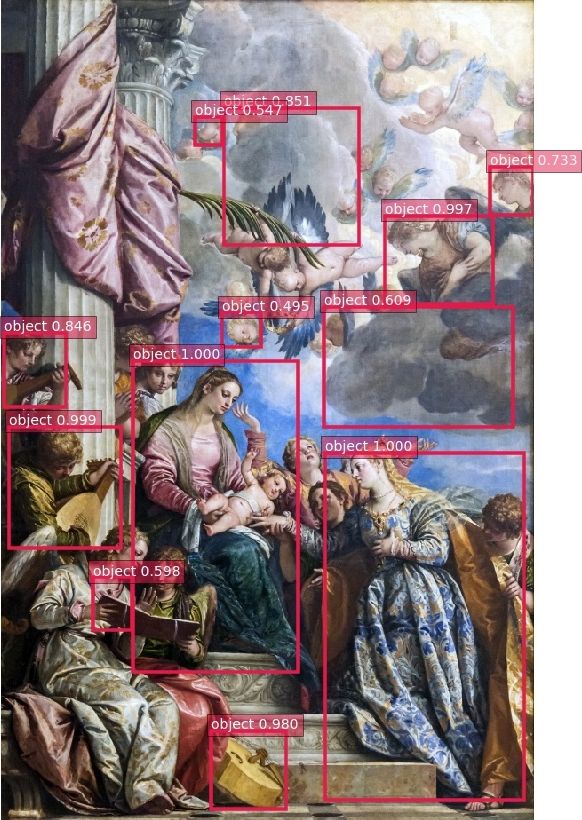}\hfill
\caption{Some of the regions of interest generated by the region proposal part (RPN) of Faster R-CNN.}
\label{fig:FasterRCNNBOX}
\end{figure}

Images are resized to 600 by 1000 before applying Faster R-CNN. We only keep the 300 boxes having  best "objectness" scores (after a NMS phase), along with their high-level features\footnote{The layer $fc7$ of size $M=2048$ in the ResNet case, often called 2048-D.}. 
An example of extracted boxes is shown in figure \ref{fig:FasterRCNNBOX}. %
About 5 images per second can be obtained on a standard GPU. This part can be performed offline since we don't fine-tune the network. 

As mentioned in \cite{kornblith_better_2018}, residual network (ResNet) appears to be the best architecture for transfer learning by feature extractions among the different ImageNet models, and we therefore choose these networks for our Faster R-CNN versions. One of them (denoted RES-101-VOC07) is a 101 layers ResNet trained for the detection task on PASCAL VOC2007. 
The other one (denoted RES-152-COCO) is a 152 layers ResNet trained on MS COCO~\cite{lin_microsoft_2014}. We will also compare our approach to the plain application of these networks for the detection tasks when possible, that is when they were trained on classes we want to detect. We refer to these approaches as FSD (fully supervised detection) in our experiments. 

For implementation, we build on the Tensorflow\footnote{\url{https://www.tensorflow.org/}} implementation of Faster R-CNN of Chen and al. \cite{chen_implementation_2017}\footnote{Code can be found on GitHub \url{https://github.com/endernewton/tf-faster-rcnn}.}.

\paragraph{\bf \MIL{}} When a new class is to be learned, the user provides a set of weakly annotated images. The \MIL{} framework described  above is then run to find a linear separator specific to the class. Note that both the database and the library of classifiers can be enriched very easily. Indeed, adding an image to the database only requires running it through the Faster R-CNN network and adding a new class only requires a MIL training.  

 For training the \MIL{}, we use a batch size of 1000 examples (for smaller sets, all features are loaded into the GPU), 300 iterations of gradient descent with a learning rate of 0.01 and  $\epsilon=0.01$ \eqref{eqn:MILS}. The whole process takes 750s for 20 classes on PASCAL VOC07 trainval (5011 images) with 12 random start points per class, on a consumer GPU (GTX 1080Ti). Actually the random restarts are performed in parallel to take advantage of the presence of the features in the GPU memory since the transfer of data from central RAM to the GPU memory is a bottleneck for our method. The 20 classes can be learned in parallel. 
 
 For the experiments of Section~\ref{sec:iconart}, we also perform a grid search on the hyper-parameter C \eqref{eqn:LossReg} by splitting the training set into training and validation sets. We learn several couples ($w$,$b$) for each possible value of C (different initialisation) and the one that minimises the loss \eqref{eqn:MILS} for each class is selected.

\section{Experiments}
\label{sec:experiments}

In this section, we perform weakly supervised detection experiments on different databases, in order to illustrate different assets of our approach. 

In all cases, and besides other comparisons, we compare our approach (\MIL{}) to the following baseline, which is actually the approach chosen for the detection experiments in~\cite{crowley_art_2016} (except that we do not perform box expansion): the idea is to consider that the region with the best "objectness" score is the region corresponding to the label associated to the image (positive or negative). This baseline will be denoted as \MAX{}. Linear-SVM classifier are learnt using those features per class in a one-vs-the-rest manner. The weight parameter that produces the highest AP (Average Precision) score is selected for each class by a cross validation method\footnote{We use a 3-fold cross validation while \cite{crowley_art_2016} use constant training and validation set.} and then a classifier is retrained with the best hyper-parameter on all the training data per class. This baseline requires to train several SVMs and is therefore costly.

At test time, the labels and the bounding boxes are used to evaluate the performance of the methods in term of AP par class. The generated boxes are filtered by a NMS with an Intersection over Union (IoU) \cite{everingham_pascal_2007} threshold of 0.3 and a confidence threshold of 0.05 for all methods.

\subsection{Experiments on PASCAL VOC}
Before proceeding with the transfer learning and testing our method on paintings, we start with a sanity check experiment on PASCAL VOC2007~\cite{everingham_pascal_2007}. We compare our weakly supervised approach, \MIL{}, to the plain application of the fully supervised Faster R-CNN~\cite{ren_faster_2015} and to the weakly supervised MAX procedure recalled above. We perform the comparison using two different architectures (for the three methods), RES-101-VOC07 and RES-512-COCO, as explained in the previous section. 

\begin{savenotes}
 \begin{table}
 \caption{ \textbf{VOC 2007 test} Average precision (\%). Comparison of the Faster R-CNN detector (trained in a fully supervised manner : FSD) and our \MILS{} algorithm (trained in a weakly supervised manner) for two networks RES-101-VOC07 and RES-152-COCO.}
\label{table:VOC2007}
 \resizebox{\columnwidth}{!}{
\begin{tabular}{|c|c|cccccccccccccccccccc|c|}
 Net & Method & aero & bicy & bird & boa & bot & bus & car & cat & cha & cow & dtab & dog  & hors & mbik  & pers &  plnt &  she & sofa & trai & tv & mean \\  \hline \hline
 RES- & FSD \cite{he_deep_2015} & 73.6 & 82.3 & 75.4 & 64.0 & 57.4 & 80.2 & 86.5 & 86.2 & 52.7 & 85.2 & 66.9 & 87.0 & 87.1 & 82.9 & 81.2 & 45.7 & 76.8 & 71.2 & 82.6 & 75.5 & 75.0 \\
  101- & \MAX{} & 20.8 & 47.0 & 26.1 & 20.2 & 8.3 & 41.1 & 44.9 & 60.1 & 31.7 & 54.8 & 46.4 & 42.9 & 62.2 & 58.7 & 20.9 & 21.6 & 37.6 & 16.7 & 42.0 & 19.8 & 36.2 \\ 
 VOC07  & \MILS{}\footnote{It is the average performance on 100 runs of our algorithm.}   & 63.5  & 78.4   & 68.5   & 54.0  & 50.7  & 71.8  & 85.6  & 77.1   & 52.7  & 80.0   & 60.1  & 78.3   & 80.5  & 73.5  & 74.7  & 37.4   & 71.2  & 65.2  & 75.7   & 67.7  & 68.3   $\pm$ 0.2   \\ 
   \hline
 RES- & FSD \cite{he_deep_2015}  &  91.0 & 90.4 & 88.3 & 61.2 & 77.7 & 92.2 & 82.2 & 93.2 & 67.0 & 89.4 & 65.8 & 88.0 & 92.0 & 89.5 & 88.5 & 56.9 & 85.1 & 81.0 & 89.8 & 85.2 & 82.7 \\
   152- & \MAX{}~\cite{crowley_art_2016} & 58.8 & 64.7 & 52.4 & 8.6 & 20.8 & 55.2 & 66.8 & 76.1 & 19.4 & 66.3 & 6.7 & 59.7 & 56.4 & 43.3 & 15.5 & 18.3 & 80.3 & 7.6 & 71.8 & 32.6 & 44.1 \\
 COCO  & \MILS{}\repeatfootnote &   88.0  & 90.2   & 84.3   & 66.0 & 78.7   & 93.8   & 92.7   & 90.7   & 63.7  & 78.8  & 61.5  & 88.4  & 90.9  & 88.8  & 87.9   & 56.8  & 75.5  & 81.3   & 88.4  & 86.1  & 81.6 $\pm$ 0.3 \\ 
\end{tabular}
}
\end{table}
\end{savenotes}

As shown in Table \ref{table:VOC2007} our weakly supervised approach (only considering annotations at the image level~\footnote{However, observe that since we are relying on Faster R-CNN, our system uses a subpart trained using class agnostic bounding boxes.}) 
yields performances that are only slightly below the ones of the fully supervised approach (using instance-level annotations).
On the average, the loss is only 1.1\% of mAP when using RES-512-COCO (for both methods). The baseline \MAX{} procedure (used for transfer learning on paintings in~\cite{crowley_search_2014}) yields notably inferior performances. 

\subsection{Detection evaluation on Watercolor2k and People-Art databases}

We compare our approach with two recent methods performing object detection in artworks, one in a fully supervised way~\cite{westlake_detecting_2016} for detecting people, the other using a (partly) weakly supervised method to detect several VOC classes on watercolor images~\cite{inoue_crossdomain_2018}. For the learning stage, the first approach uses instance-level annotations on paintings, while the second one uses instance-level annotations on photographs and image-level annotations on paintings. In both cases, it is shown that using image-level annotations only (our approach, \MIL{}) only yields  a light loss of performances. 

\paragraph{\bf Experiment 1 : Watercolor2k} This database, introduced in \cite{inoue_crossdomain_2018}, and available online\footnote{\url{https://github.com/naoto0804/cross-domain-detection}}, is a subset of watercolor artworks from the \textbf{BAM!} database \cite{wilber_bam_2017}  with instance-level annotations for 6 classes (bike, bird, dog, cat, car, person) that are included in the PASCAL VOC, in order to study cross-domain transfer learning.
On this database, we compare our approach to the methods from~\cite{inoue_crossdomain_2018} and from~\cite{bilen_weakly_2016}, to the baseline \MAX{} discussed above, as well as to the classical MIL approach MI-SVM~\cite{andrews_support_2003} (using a maximum of 50 iterations and no restarts). 

In \cite{inoue_crossdomain_2018}, a style transfer transformation (Cycle-GAN \cite{zhu2017a}) is applied to natural images with instance-level annotation. The images are transferred to the new modality (i.e. watercolor) in order to fine-tune a detector pre-trained on natural images. This detector is used to predict localisation of objects on watercolor images annotated at the image level. The detector is then fine-tuned on those images in a fully supervised manner. Bilen and Vedaldi \cite{bilen_weakly_2016} proposed a Weakly Supervised Deep Detection Network (WSDDN), which consists in transforming a pre-trained network by replacing its classification part by a two streams network (a region proposal stream and a classification one) combined with a weighted MIL pooling strategy.

\begin{savenotes}
\begin{table}
\caption{\textbf{Watercolor2k (test set)} Average precision (\%). Comparison of the proposed \MILS{} method to alternative approaches.}
\label{table:Watercolor2k}
 \resizebox{\columnwidth}{!}{
\begin{tabular}{|c|c|cccccc|c|}
 Net & Method  & bike &bird &car &cat &dog &person & mean \\  \hline \hline
VGG & WSDDN \cite{bilen_weakly_2016} \footnote{The performance come from the original paper \cite{inoue_crossdomain_2018}.}  &  1.5 & 26.0 & 14.6 & 0.4 & 0.5 & 33.3 & 12.7 \\
SSD & DT+PL \cite{inoue_crossdomain_2018}  \repeatfootnote & 76.5 & 54.9 & 46.0 & 37.4 & 38.5 & 72.3 & 54.3 \\
  \hline %
  \multirow{2}{*}{RES-152-COCO}   & \MAX{}~\cite{crowley_art_2016} & 74.0 & 34.5 & 26.8 & 17.8 & 21.5 & 21.0 & 32.6 \\
 & MI-SVM \cite{andrews_support_2003} & 66.8 & 23.5 & 6.7 & 13.0 & 8.4 & 14.1 & 22.1 \\
\hline \hline
    & \MILS{} [Our] \footnote{Standard deviation computed on 100 runs of the algorithm.}  & 85.2  & 48.2 & 49.2   & 31.0   & 30.0   & 57.0  & 50.1   $\pm$ 1.1  \\  %
\end{tabular}}

\end{table}
\end{savenotes}

From Table~\ref{table:Watercolor2k}, one can see that our approach performs clearly better than the other ones using image-level annotations only (\cite{bilen_weakly_2016}, \MAX{}, MI-SVM). We also observe only a minor degradation of average performances  (54.3 \% versus 48.9 \%) with respect to the method~\cite{inoue_crossdomain_2018}, which is retrained using style transfer and instance-level annotations on photographs. 

\paragraph{\bf Experiment 2 : People-Art} This database, introduced in \cite{westlake_detecting_2016}, is made of artistic images and bounding boxes for the single class {\it person}. This database is particularly challenging because of its high variability in styles and depiction techniques. The method introduced in~\cite{westlake_detecting_2016} yields excellent detection performances on this database, but necessitates instance-level annotations for training. The authors rely on Fast R-CNN \cite{girshick_fast_2015}, of which they only keep the three first layers, before re-training the remaining of the network using manual location annotations on their database.

In Table~\ref{tab:People-Art_detection}, one can see that our approach MI-max yields detection results that are very close to the fully supervised results from~\cite{westlake_detecting_2016}, despite a much lighter training procedure. In particular, as already explained, our procedure can be trained directly on large, globally annotated database, for which manually entering instance-level annotations is tedious and time-costly. 
\begin{savenotes}
\begin{table}
\centering
\caption{\textbf{People-Art (test set)} Average precision (\%). Comparison of the proposed \MILS{} method to alternative approaches.}
\label{tab:People-Art_detection}
\begin{tabular}{|c|c|c|}
Net & Method & person \\
\hline
\hline 
Fast R-CNN (VGG16)  & Fine tuned \cite{westlake_detecting_2016}\footnote{The performance come from the original paper.} &  59 \\
\hline
\multirow{2}{*}{RES-152-COCO}  & \MAX{}~\cite{crowley_art_2016}  & 25.9 \\
& MI-SVM \cite{andrews_support_2003} & 13.3  \\
\hline \hline
 RES-152-COCO & \MILS{}   [Our]  & 55.4  $\pm$ 0.7 \\
\end{tabular}

 \end{table}  
 \end{savenotes}

\subsection{Detection on \NB{} database}
\label{sec:iconart}

In this last experimental section, we investigate the ability of our approach to learn and detect new classes that are specific to the analysis of artworks, some of which cannot be learnt on photographs. Typical such examples include iconic characters in certain situations, such as Child Jesus, the crucifixion of Jesus, Saint Sebastian, etc.
Although there has been a recent effort to increase open-access  databases of artworks by academia and/or museums workforce \cite{mensink_rijksmuseum_2014,crowley_search_2014,lecoutre_recognizing_2017,mao_deepart_2017,strezoski_omniart_2017,rijksmuseum_online_2018,europeana_collections_2018,met_image_2018}, they usually don't include systematic and reliable keywords. One exception is  the database from the Rijkmuseum, with labels based on the IconClass classification system \cite{iconclass_home_2018}, but this database is mostly composed of prints, photographs and drawings. Moreover, these databases don't include the localisation of objects or characters. 

\begin{figure}
     \centering
     \hfill
         \includegraphics[height=\heightimage]{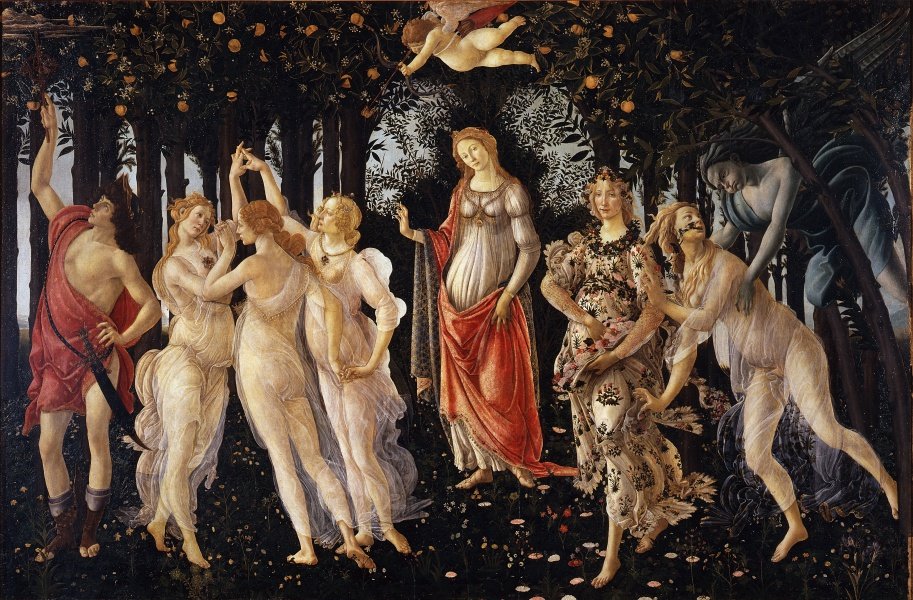} \hfill
         \includegraphics[height=\heightimage]{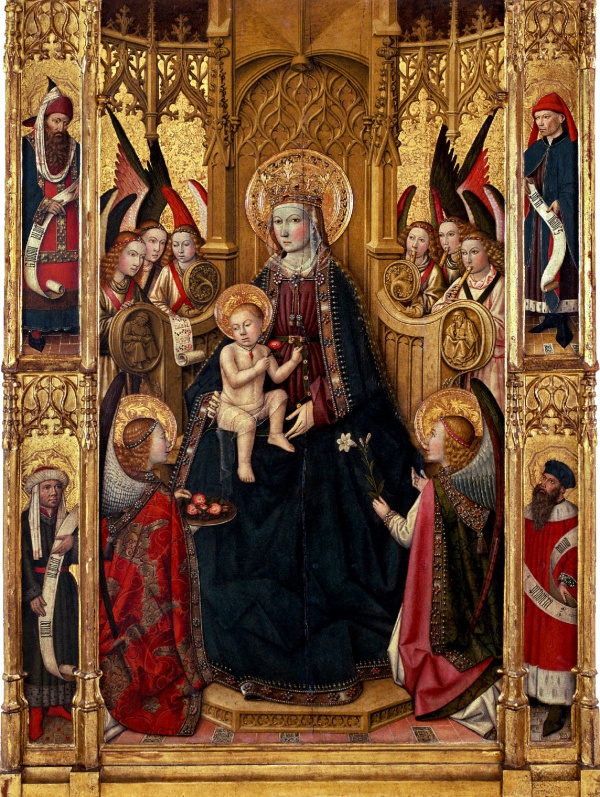} \hfill
     \includegraphics[height=\heightimage]{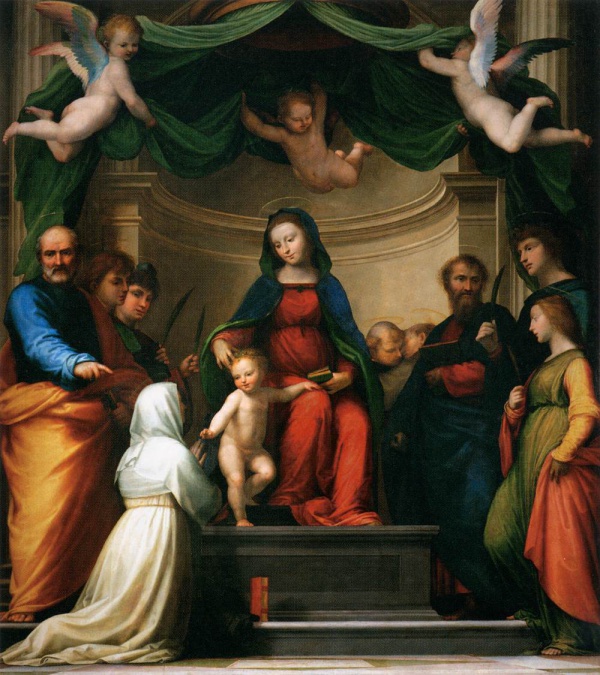} \hfill
     \includegraphics[height=\heightimage]{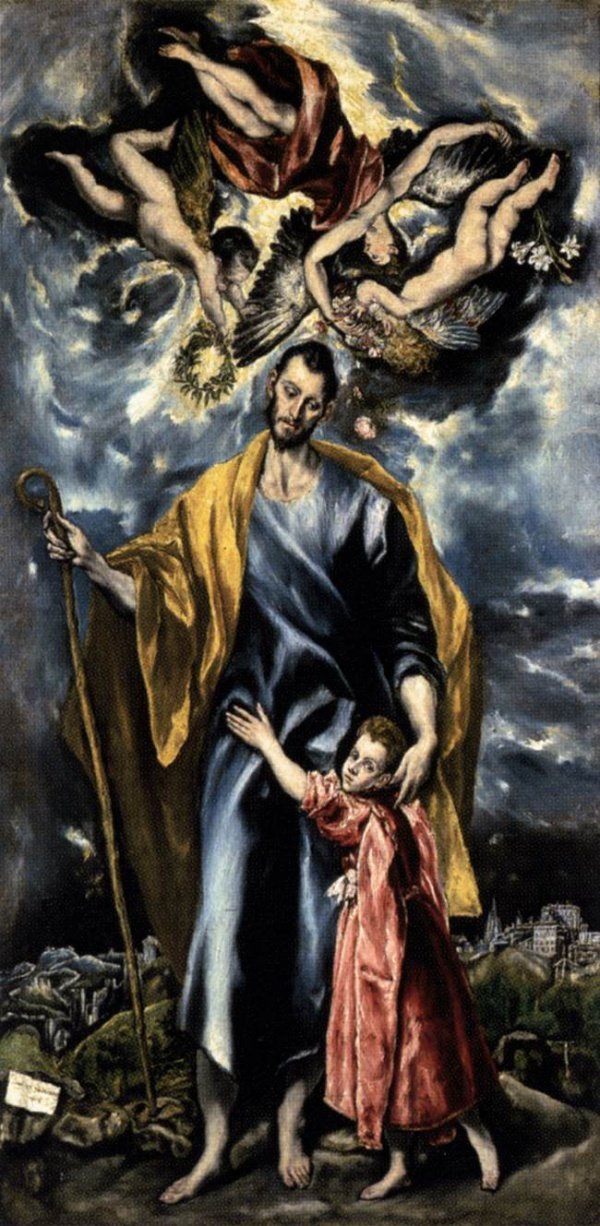} \hfill
     \includegraphics[height=\heightimage]{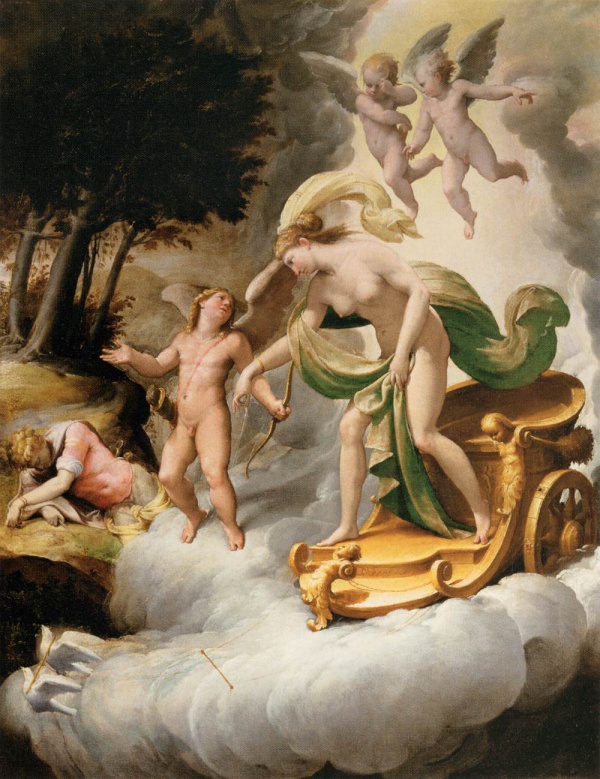} \hfill \\
     \centering
  \hfill
     \includegraphics[height=\heightimage]{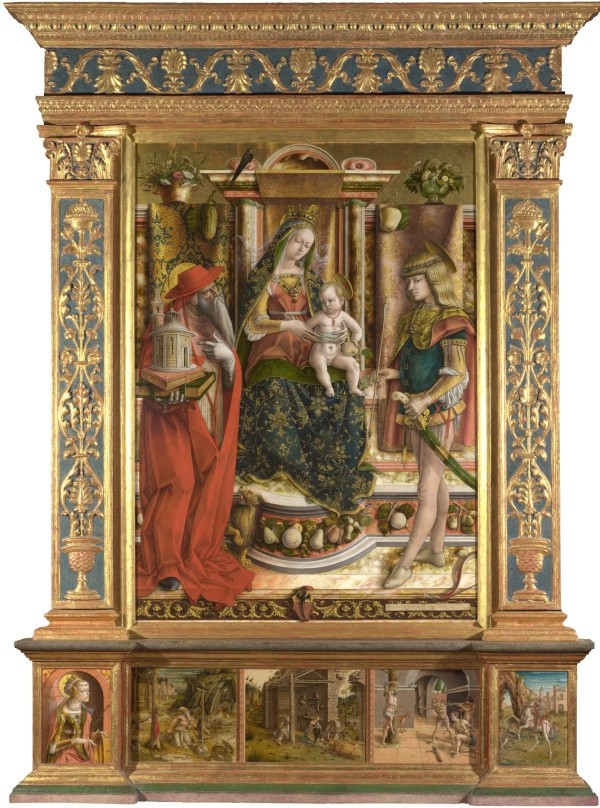} \hfill \includegraphics[height=\heightimage]{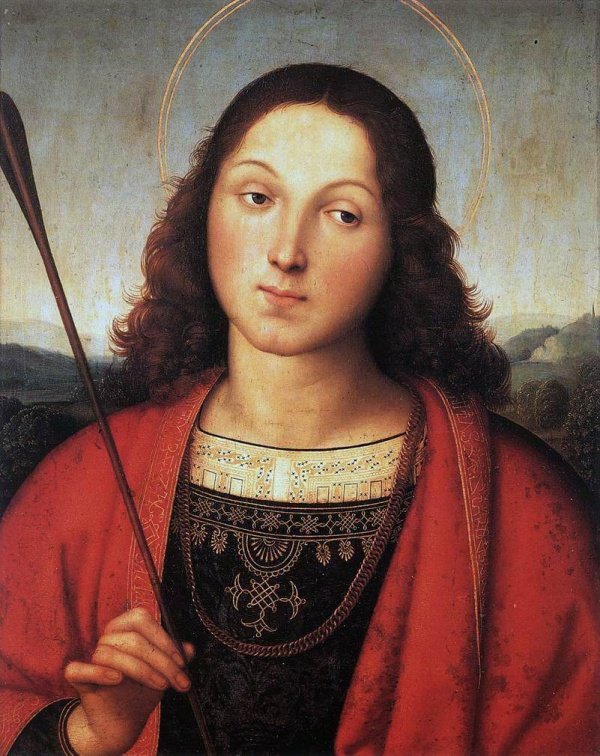} \hfill
     \includegraphics[height=\heightimage]{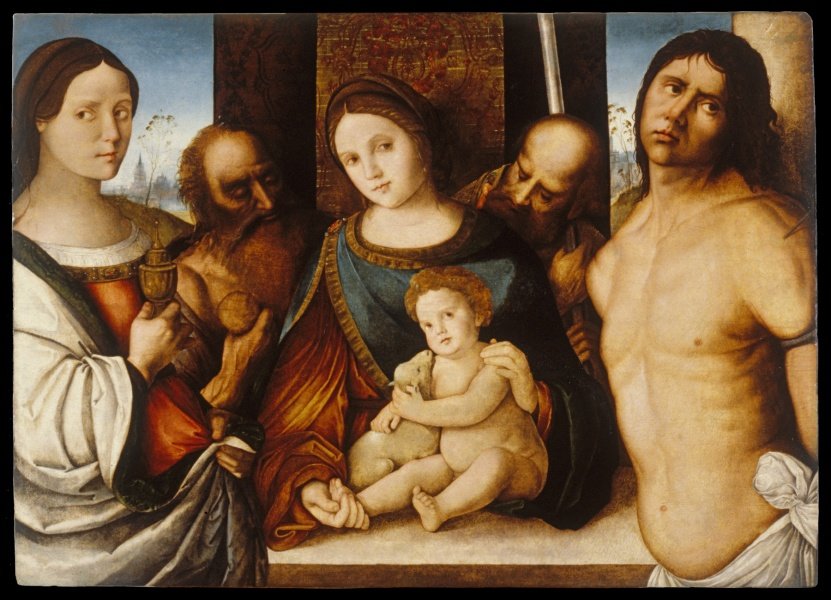} \hfill
     \includegraphics[height=\heightimage]{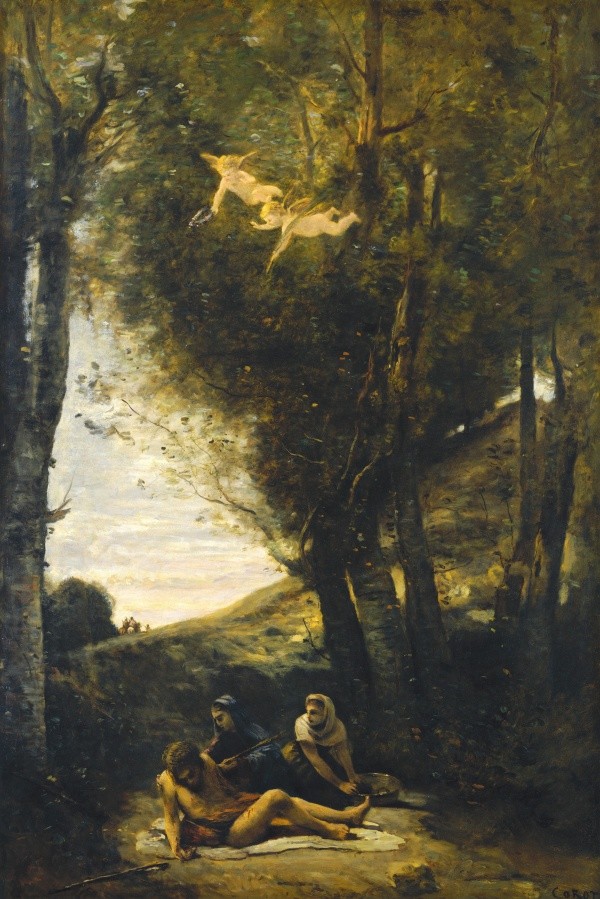} \hfill
    \includegraphics[height=\heightimage]{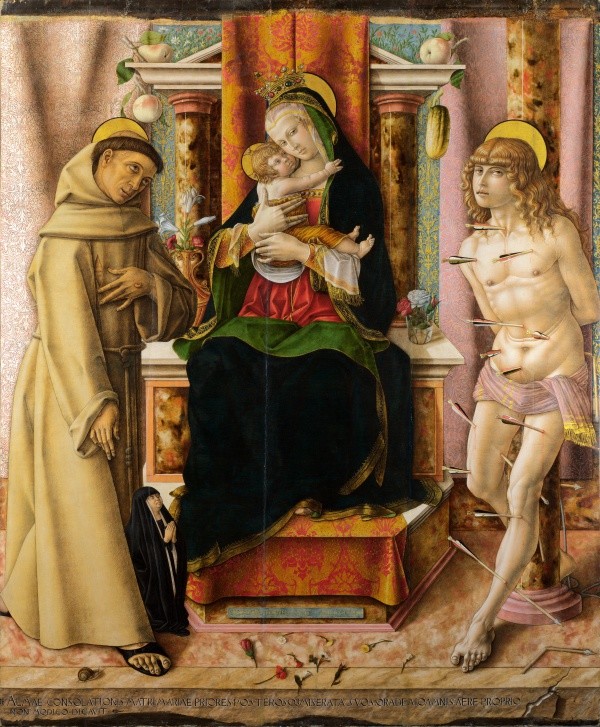} \hfill \\
     \centering
    \caption{Example images from the \NB{} database. Angel on the first line, Saint Sebastian on the second. We can see some of the challenges posed by this database: tiny objects, occlusions and large pose variability.}
    \label{fig:NotreBase}
\end{figure}

In order to study the ability of our (and other) systems to detect iconographic elements, we gathered 5955 painting images from Wikicommons\footnote{\url{https://commons.wikimedia.org/wiki/Main_Page}}, ranging from the 11th to the 20th century, which are partially annotated by the Wikidata\footnote{\url{https://www.wikidata.org/wiki/Wikidata:Main_Page}} contributors. We manually checked and completed image-level annotations for 7 classes. The dataset is split in training and test sets, as shown in Table \ref{tab:WikiDataSet}.
For a subset of the test set, and only for the purpose of performance evaluation,  instance-level annotations have been added. The resulting database is called \NB{}\footnote{The database is available online \url{https://wsoda.telecom-paristech.fr/downloads/dataset/IconArt_v1.zip}.}. Example images are shown in Figure  \ref{fig:NotreBase}. To the best of our knowledge, the presented experiments are the first investigating the ability of modern detection tools to classify and detect such iconographic elements in paintings. Moreover, we investigate this aspect in a weakly supervised manner.

\begin{table}
    \centering
    \resizebox{\columnwidth}{!}{
   \begin{tabular}{|c|c|c|c|c|c|c|c|c|c|}
\hline 
Class & Angel & Child Jesus & Crucifixion &  Mary & nudity & ruins & Saint Sebastian & None & Total \\ 
\hline 
Train & 600 & 755 & 86 & 1065  &  956 & 234 & 75 & 947 & 2978 \\
\hline 
Test for classification & 627 & 750 & 107 & 1086 & 1007 &  264 & 82 & 924 & 2977 \\
Test for detection  & 261 & 313 & 107 & 446 & 403 &  114 & 82 & 623 & 1480 \\
\hline 
Number of instances & 1043 &320 & 109 & 502 & 759 & 194 & 82 & & 3009 \\
\hline
\end{tabular}}
    \caption{Statistics of the IconArt database}
    \label{tab:WikiDataSet}
    
\end{table}

To fix ideas on the difficulty of dealing with iconographic elements, we start with a classification experiment. For this, we use the same classification approach as in~\cite{crowley_search_2014}, using InceptionResNetv2 \cite{szegedy_inceptionv4_2016} as a feature extractor\footnote{Only the center of the image is provided to the network and extracted features are 1536-D.}. We also perform classification-by-detection experiments, using the previously described \MAX{} approach (as in~\cite{crowley_art_2016}) and our approach, \MILS{}. In both cases, for each class, the score at the image level is the highest confidence detection score for this class on all the regions of the image. Results are displayed in Table~\ref{tab:WikiTenLabels_classif}. First, we observe that classification results are very variable depending on the class. Classes such as Jesus Child, Mary or crucifixion have relatively high classification scores. Others, such as Saint Sebastian, are only scarcely classified, probably due to a limited quantity of examples and a high variability of poses, scales and depiction styles. We can also observe that, as mentioned in \cite{crowley_art_2016}, the classification by detection can provide better scores than global classification, possibly because of small objects, such as angels in our case. Observe that these classification scores can probably be increased using multi-scale learning (as in \cite{vannoord_learning_2017}), augmentation schemes and an ensemble of networks \cite{crowley_art_2016}.  

\begin{table}
\caption{\textbf{\NB{} classification test set} \textit{classification} average precision (\%).}
\label{tab:WikiTenLabels_classif}
 \resizebox{\columnwidth}{!}{
\begin{tabular}{|c|c|ccccccc|c|}
Net & Method &angel & JCchild & crucifixion & Mary & nudity & ruins & StSeb & mean \\ \hline
\hline 
InceptionResNetv2 \cite{szegedy_inceptionv4_2016}  &  & 44.1 & 77.2 &  57.8 & 81.1 &  77.4 &  74.6 & 26.8 &  62.7 \\
\hline
& \MAX{}~\cite{crowley_art_2016}  & 49.3 &  74.7 & 30.3 & 67.5 & 57.4 & 43.2 & 7.0 &  47.1 \\ 
RES-152-COCO & \MILS{} [Our]  &  57.4   & 60.7  & 79.9  & 70.4   & 65.3   & 45.9   & 17.0  & 56.7   $\pm$ 1.0   \\ 
& \MILSCV{} [Our]  & 61.0   & 68.9   & 80.2  & 71.4  & 66.3   & 51.7 & 14.8   & 59.2   $\pm$ 1.2
\end{tabular}}

 \end{table}

Next, we evaluate the detection performance of our method, first with a restrictive metric : AP per class with an IoU $\geqslant$0.5 (as in all previous detection experiments in this paper), then with a less restrictive metric with IoU $\geqslant$0.1. Results are displayed in Table \ref{tab:WikiTenLabels_detection}. Results on this very demanding experiment are a mixed-bag. Some classes, such as crucifixion, and to a less extend nudity or Jesus Child are correctly detected. Others, such as angel, ruins or Saint Sebastian, hardly get it up to 15\% detection scores, even when using the relaxed criterion IoU $\geqslant$0.1. Beyond a relatively small number of examples and very strong scale and pose variations, there are further reasons for this :
\begin{itemize}
    \item The high in-class depiction variability (for Saint Sebastian for instance) 
    \item The many occlusions between several instances of a same class (angel)
    \item The fact that some parts of an object can be more discriminative than the whole object (nudity)
\end{itemize}{}
 
 \begin{table}
\caption{\textbf{\NB{} detection test set} \textit{detection} average precision (\%). All methods based on RES-152-COCO.}
\label{tab:WikiTenLabels_detection}
 \resizebox{\columnwidth}{!}{
\begin{tabular}{|c|c|ccccccc|c|}
 Method & Metric &angel & JCchild & crucifixion & Mary & nudity & ruins & StSeb & mean \\ \hline
 \hline
\multirow{2}{*}{\MAX{}~\cite{crowley_art_2016}}  & AP IoU $\geqslant$0.5 & 1.4 &  3.9 & 7.4 & 2.8 & 3.9 & 0.3 & 0.9 & 2.9 \\ 
 & AP IoU $\geqslant$0.1 & 10.1 & 36.2 & 28.2 & 18.4 & 14.0 & 1.6 & 2.8 &  15.9 \\ 
 \hline
 \hline
 \multirow{2}{*}{\MILS{} [Our]} &  AP IoU $\geqslant$0.5 & 0.3  & 0.9  & 37.3  & 3.8   & 21.2   & 0.5  & 10.9   & 10.7   $\pm$ 1.7   \\
& AP IoU $\geqslant$0.1 & 6.4  & 25.3 & 74.4  & 44.6   & 30.9  & 6.8  & 17.2 & 29.4   $\pm$ 1.7   \\  
 \hline
 \multirow{2}{*}{\MILSCV{} [Our] } & AP IoU $\geqslant$0.5 & 3.0   & 17.7   & 32.6   & 4.8  & 23.5   & 1.1   & 9.6   & 13.2   $\pm$ 3.1   \\ 
 & AP IoU $\geqslant$0.1  & 12.3   & 41.2   & 74.4  & 46.3   & 31.2   & 13.6   & 16.1   & 33.6   $\pm$ 2.2   \\
\end{tabular}}
 \end{table}

\begin{figure}
\centering
  \hfill
     \includegraphics[height=\heightimage]{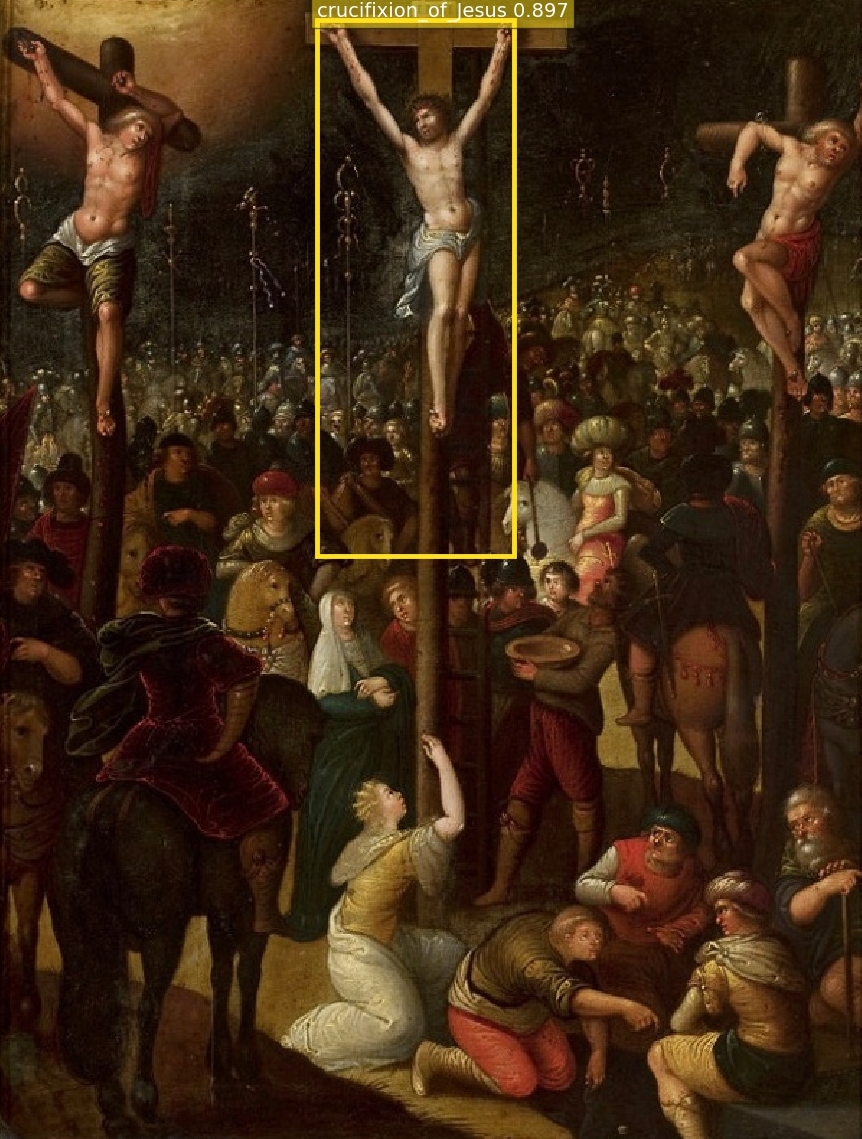}\hfill 
     \includegraphics[height=\heightimage]{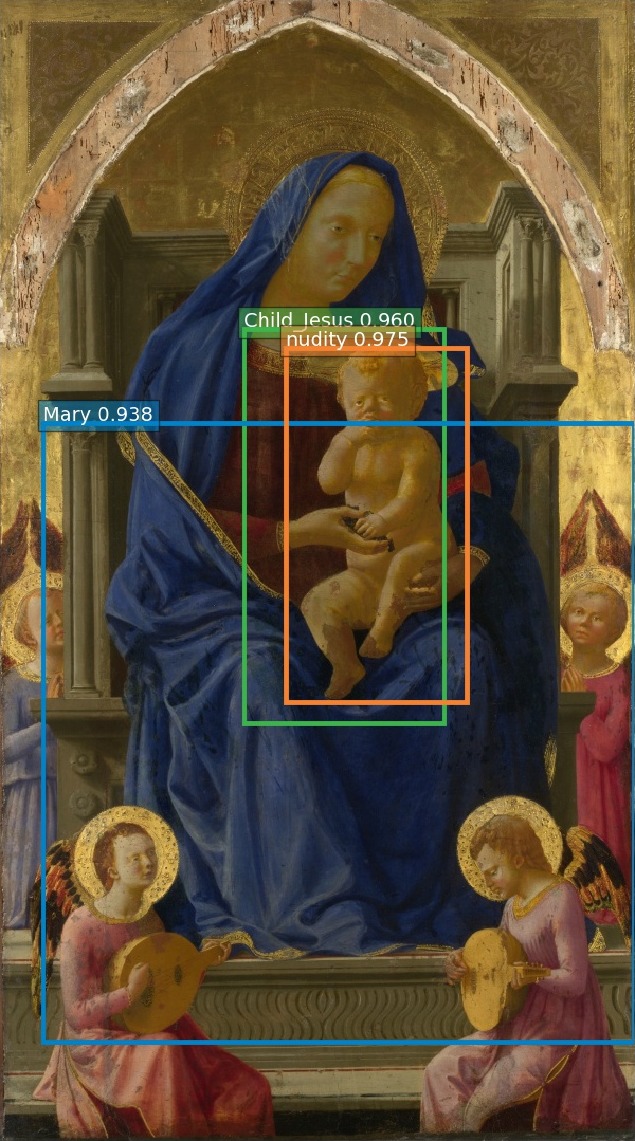} \hfill
     \includegraphics[height=\heightimage]{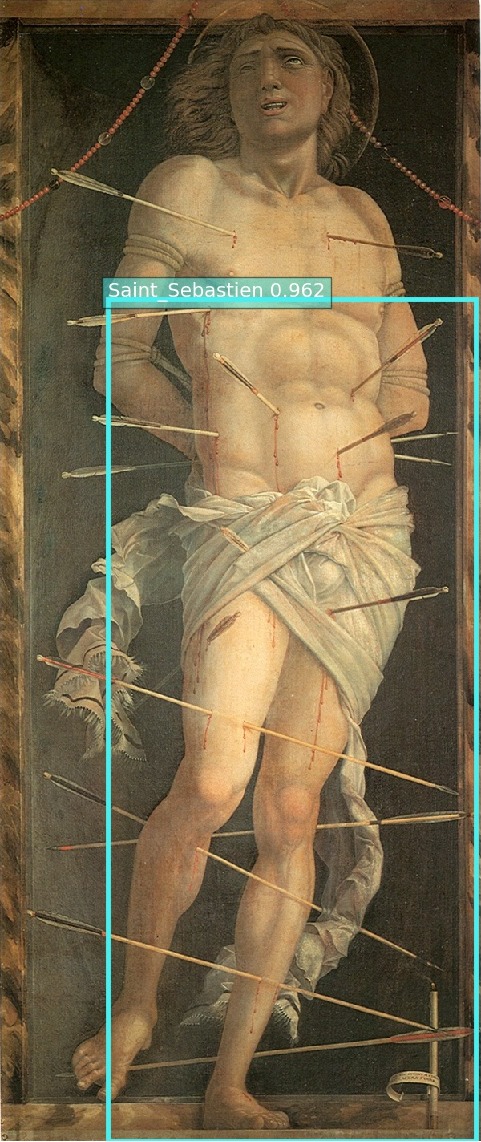} \hfill
     \includegraphics[height=\heightimage]{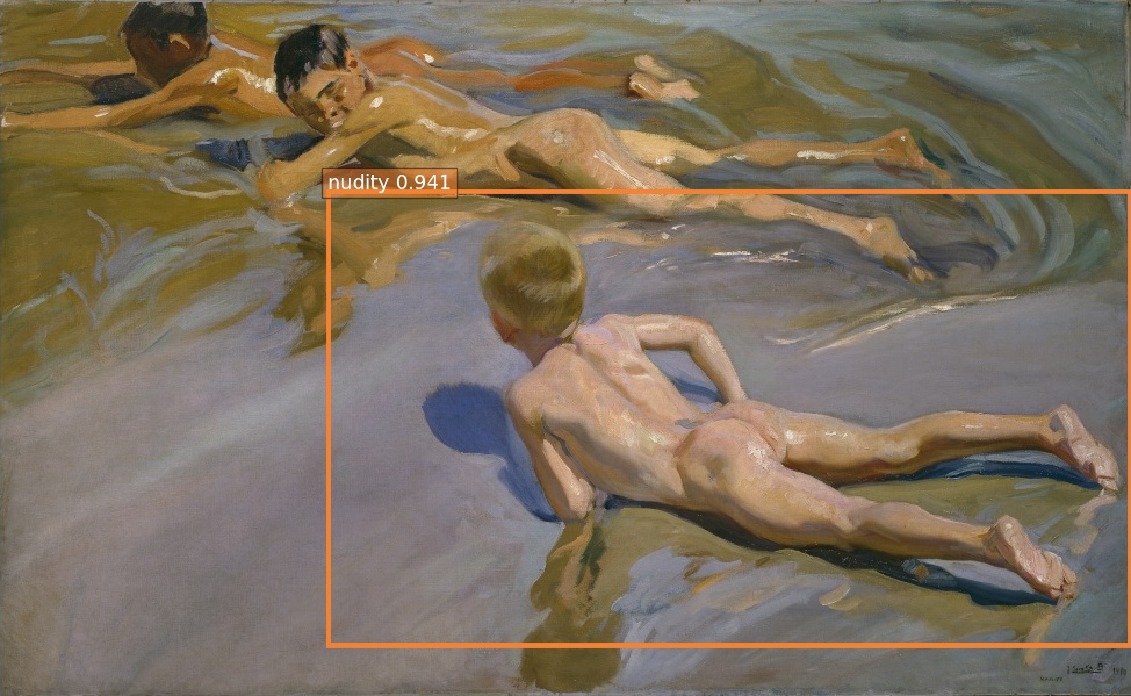} \hfill
    \includegraphics[height=\heightimage]{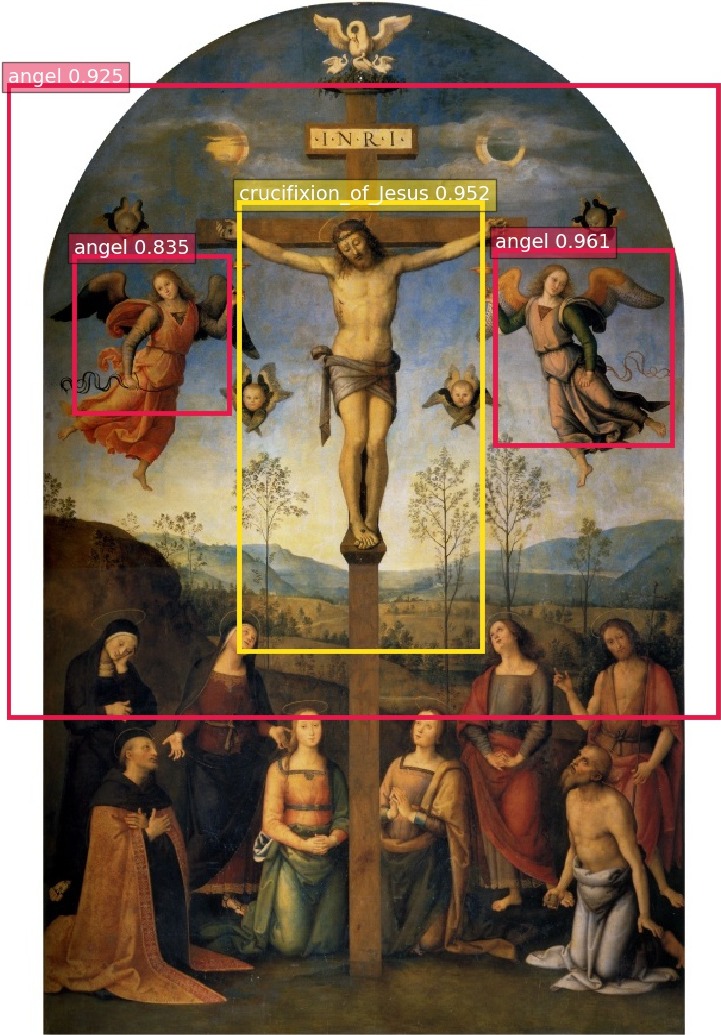} \hfill 
    \includegraphics[height=\heightimage]{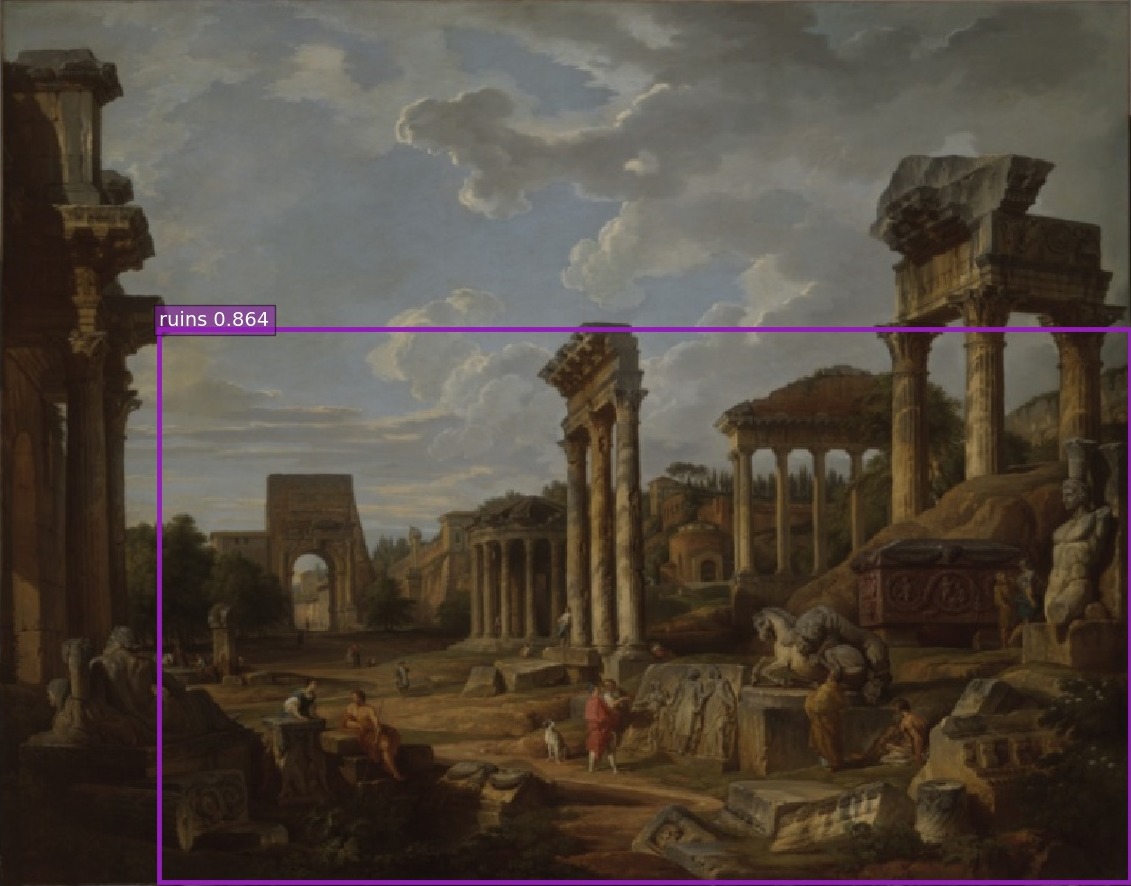} \hfill
    \includegraphics[height=\heightimage]{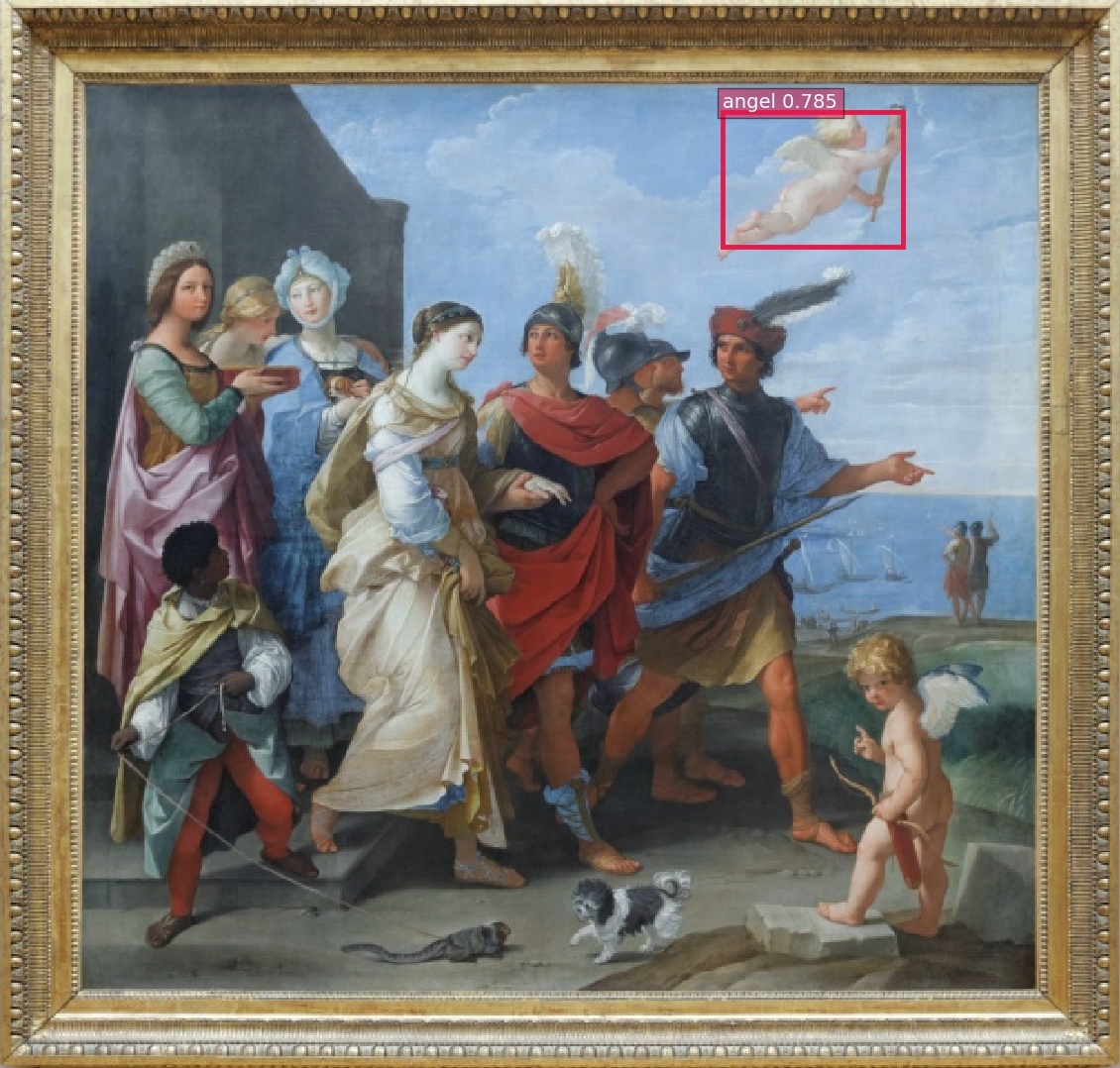} \hfill
     \includegraphics[height=\heightimage]{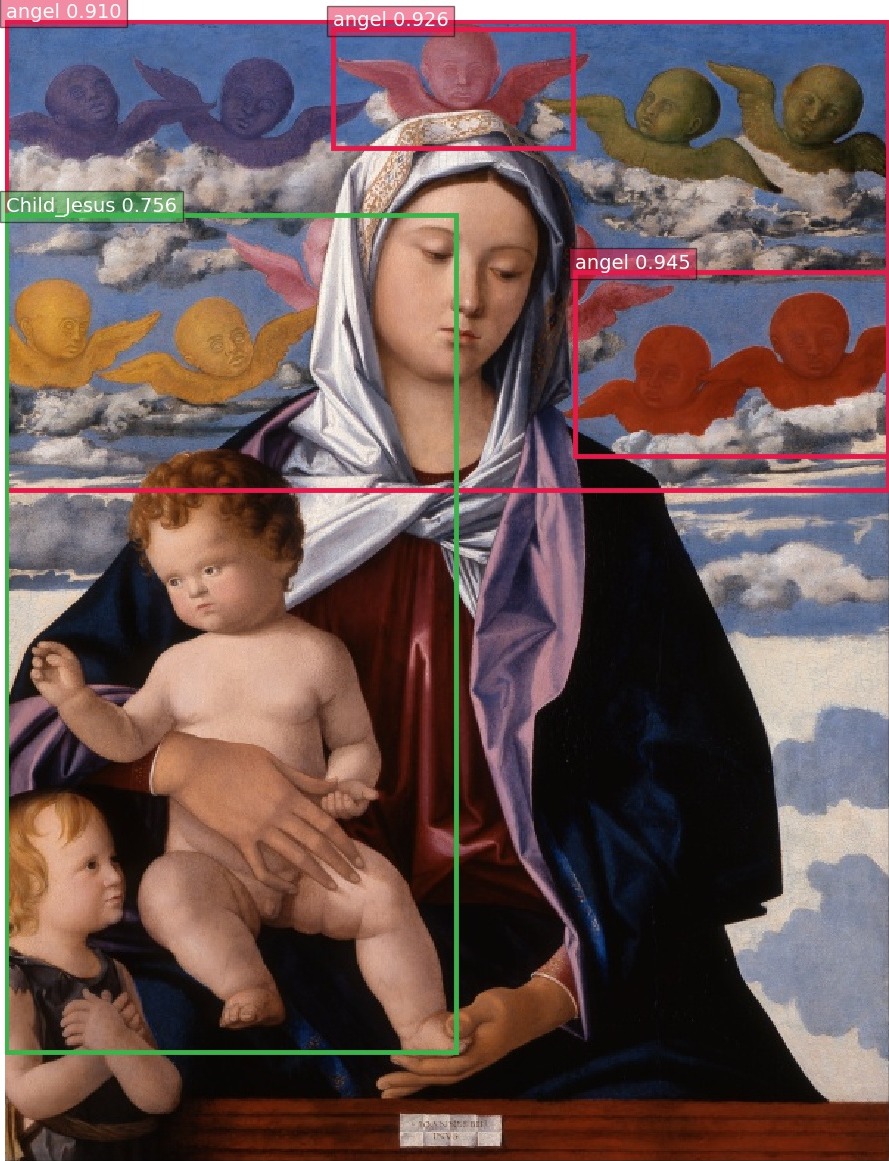} \hfill
     \includegraphics[height=\heightimage]{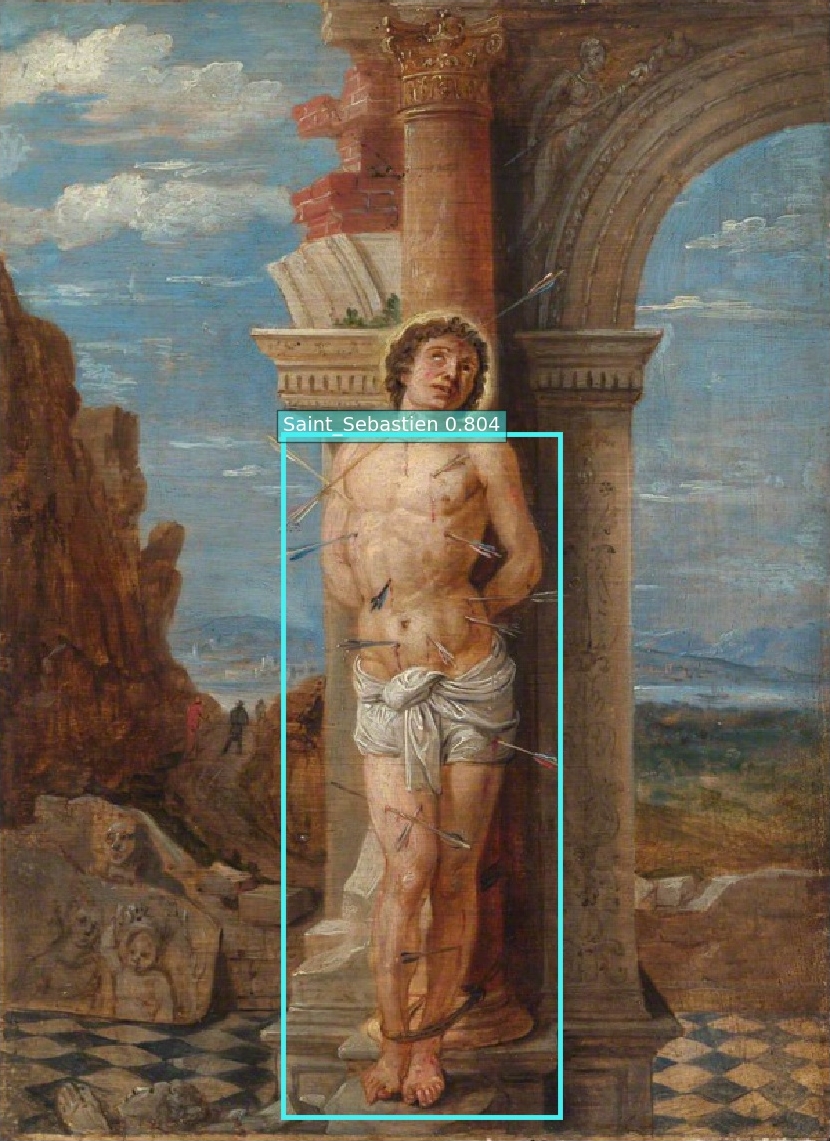} \hfill
     \includegraphics[height=\heightimage]{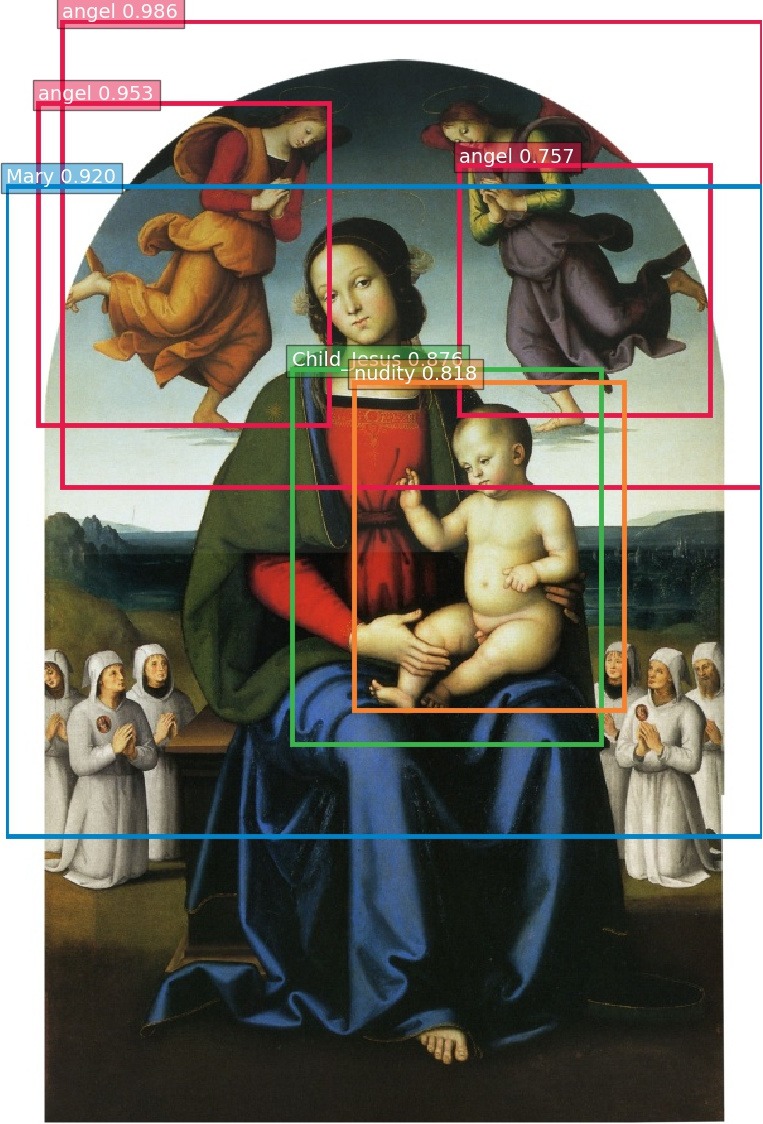} \hfill 
    \caption{Successful examples using our \MILSCV{} detection scheme. We only show boxes whose scores are over 0.75.}
    \label{fig:NotreBaseSuccesful}
\end{figure}

\begin{figure}
\centering
  \hfill
     \includegraphics[height=\heightimage]{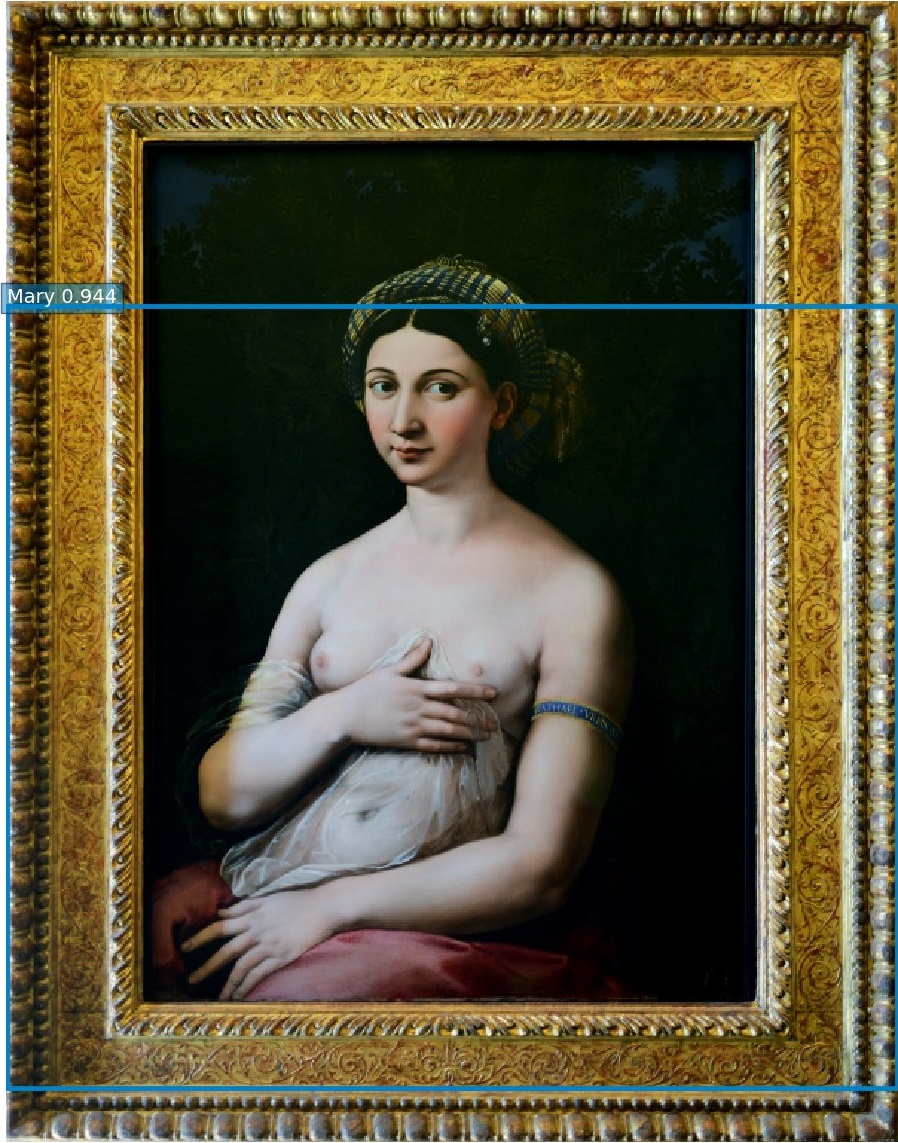} \hfill \includegraphics[height=\heightimage]{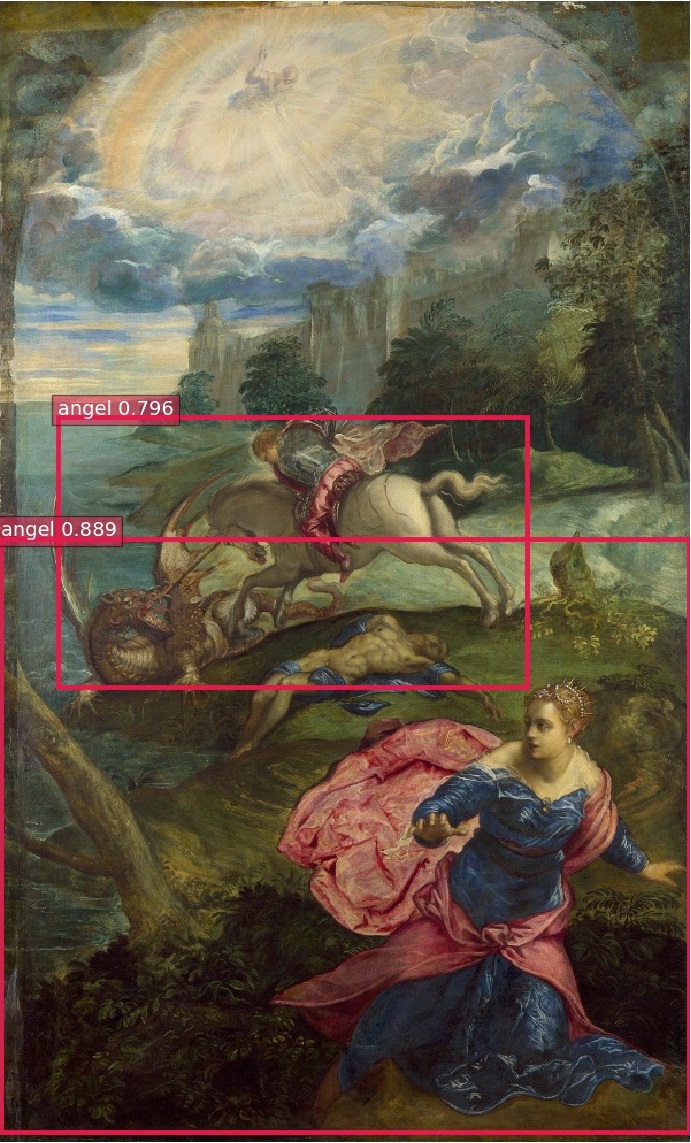} \hfill
     \includegraphics[height=\heightimage]{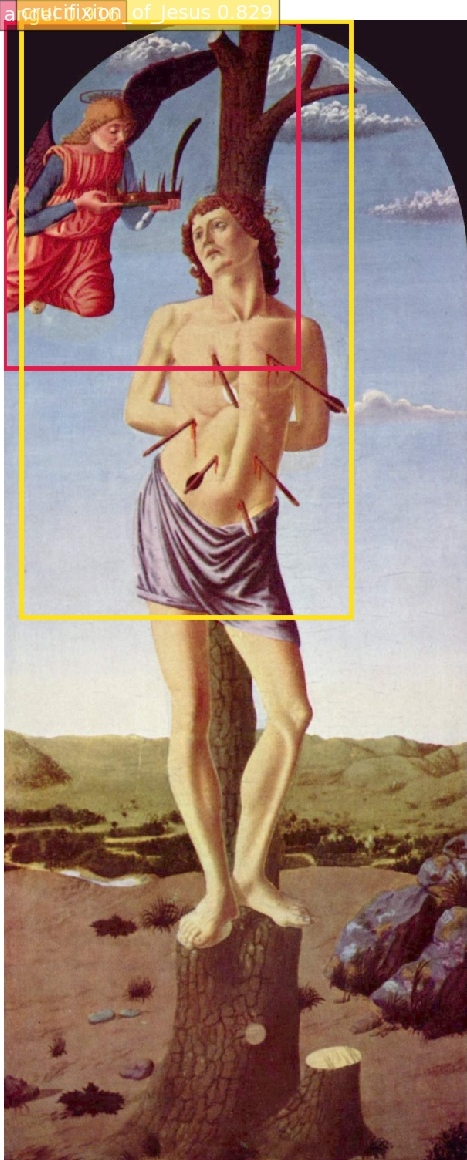} \hfill
     \includegraphics[height=\heightimage]{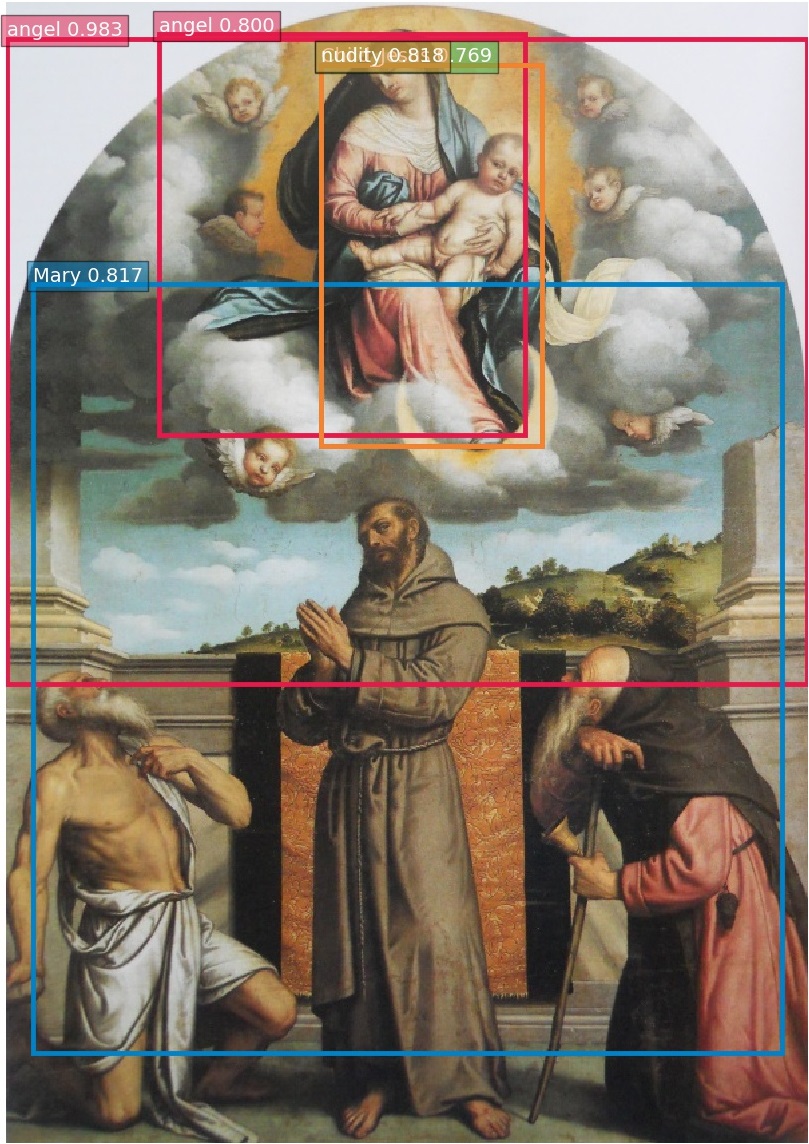} \hfill
    \includegraphics[height=\heightimage]{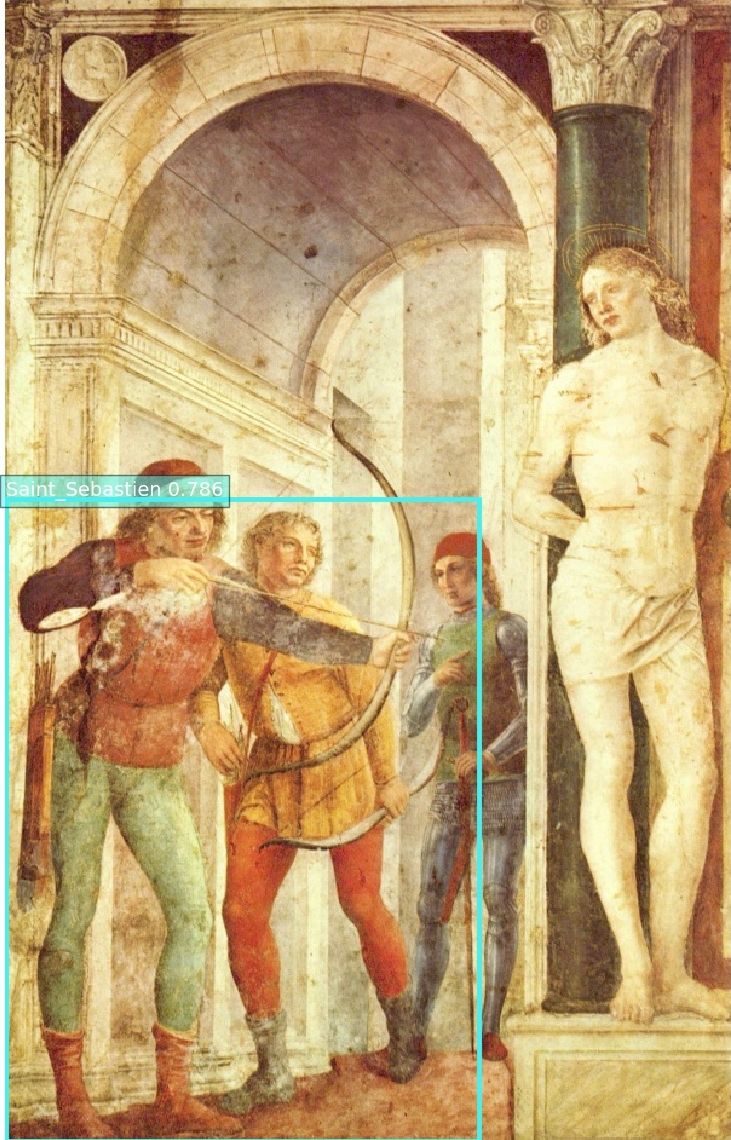} \hfill
    \caption{Failure examples using our \MILSCV{} detection scheme. We only show boxes whose scores are over 0.75.}
    \label{fig:Failures}
\end{figure}

Illustrations of successes and failures are displayed, respectively on Figures \ref{fig:NotreBaseSuccesful} and \ref{fig:Failures}. On the negative examples, one can see that often a larger region than the element of interest is selected or that a whole group of instances is selected instead of a single one.
Future work could focus on the use of several couples ($w$,$b$) instead of one to prevent those problems. 

\section{Conclusion}

Results from this paper confirm that transfer learning is of great interest to analyse artworks databases. This was previously shown for classification and fully supervised detection schemes, and was here investigated in the case of weakly supervised detection. We believe that this framework is particularly suited to develop tools helping art historians, because it avoids tedious annotations and opens the way to learning on large datasets. We also show, in this context, experiments dealing with iconographic elements that are specific to Art History and cannot be learnt on natural images. 

In future works, we plan to use localisation refinement methods, to further study how to avoid poor local optima in the optimisation procedure, to add contextual information for little objects, and possibly to fine-tune the network (as in \cite{durand_wildcat_2017}) to learn better features on artworks. Another exciting direction is to investigate the potential of weakly supervised learning on large databases with image-level annotations, such as the ones from the Rijkmuseum \cite{rijksmuseum_online_2018} or the French Museum consortium \cite{rmn_online}.

{\noindent \bf Acknowledgements.}
This work is supported by the "IDI 2017" project funded by the IDEX Paris-Saclay, ANR-11-IDEX-0003-02.

\clearpage

\bibliographystyle{splncs04}
\bibliography{VISART2018}
\end{document}